%% file: main.tex
\PassOptionsToPackage{table,xcdraw,dvipsnames}{xcolor}

\documentclass[]{fairmeta}
\input{assets/_macros}  % Add packages to _macros.tex
\usepackage{svg}
\usepackage{fontawesome5}

\newcommand{\name}{{{TV2TV}}\xspace}
\newcommand{\TtoV}{T2V\xspace}
\newcommand{\ThinktoV}{Think2V\xspace}
\newcommand{\CSGO}{CS:GO\xspace}

\newcommand{\eg}{\textit{e.g.}\xspace}

\title{\name: A Unified Framework for Interleaved Language and Video Generation}
\author[*,\dagger]{Xiaochuang Han}
\author[*,\diamond]{Youssef Emad}
\author[*,\diamond]{Melissa Hall}
\author[*,\diamond]{John Nguyen}
\author[*,\diamond]{Karthik Padthe}
\author[*,\diamond]{Liam Robbins}
\author[]{Amir Bar}
\author[]{Delong Chen}
\author[]{Michal Drozdzal}
\author[]{Maha Elbayad}
\author[]{Yushi Hu}
\author[]{Shang-Wen Li}
\author[]{Sreya Dutta Roy}
\author[]{Jakob Verbeek}
\author[]{XuDong Wang}
\author[]{Marjan Ghazvininejad}
\author[]{Luke Zettlemoyer}
\author[*]{Emily Dinan} 

\affiliation[]{Meta FAIR}

\contribution[*]{Core contributors}
\contribution[\diamond]{Ordered alphabetically}

\input{sections/00_abstract}

\correspondence{\text{Xiaochuang Han} \xspace \email{xhan77@meta.com}}
\begin{document}
\maketitle

\input{sections/01_intro}
\input{sections/03_methods}

\input{sections/04_findings}

\input{sections/05_realworld}

\input{sections/06_relatedwork}

\input{sections/07_conclusion_limitations}

\bibliographystyle{assets/plainnat}
\bibliography{main}

\newpage
\appendix
\input{sections/08_appendix}

\end{document}

%% file: assets/_macros.tex
\usepackage{tcolorbox}
\usepackage{graphicx}	
\usepackage{amsmath}	
\usepackage{amssymb}	
\usepackage{booktabs}
\usepackage{microtype}
\usepackage{epsfig}
\usepackage[table,xcdraw,dvipsnames]{xcolor}
\usepackage{caption}
\usepackage{float}
\usepackage{placeins}
\usepackage{color, colortbl}
\usepackage{stfloats}
\usepackage{enumitem}
\usepackage{tabularx}
\usepackage{xstring}
\usepackage{multirow}
\usepackage{xspace}
\usepackage{url}
\usepackage{subcaption}
\usepackage{xcolor}
\usepackage[hang,flushmargin]{footmisc}
\usepackage{amsmath}
\usepackage{graphicx}
% ...

\newcommand{\R}[1]{{%
    \textbf{%
        \ifstrequal{#1}{1}{\textcolor{red}{R#1}}{%
        \ifstrequal{#1}{2}{\textcolor{blue}{R#1}}{%
        \ifstrequal{#1}{3}{\textcolor{magenta}{R#1}}{%
        \ifstrequal{#1}{4}{\textcolor{teal}{R#1}}{%
                           \textcolor{cyan}{R#1}%
        }}}}%
    }%
}}

\newtcbox{\hlprimarytab}{on line, box align=base, colback=BlueGreen!20, size=fbox, before upper=\strut, top=-1.5pt, bottom=-1.5pt, left=-1pt, right=-1pt, boxrule=0pt}

\newtcbox{\hlsecondarytab}{on line, box align=base, colback=WildStrawberry!10,size=fbox, before upper=\strut, top=-1.5pt, bottom=-1.5pt, left=-2pt, right=-2pt, boxrule=0pt}

%% file: sections/00_abstract.tex
\abstract{
Video generation models are rapidly advancing, but can still struggle with complex video outputs that require significant semantic branching or repeated high-level reasoning about what should happen next.
In this paper, we introduce a new class of omni video-text models that integrate ideas from recent LM reasoning advances to address this challenge. 
More specifically, we present \name, a unified generative modeling framework which decomposes video generation into an interleaved text and video generation process. 
\name jointly learns language modeling (next-token prediction) and video flow matching (next-frame prediction) using a Mixture-of-Transformers (MoT) architecture.
At inference time, \name decides when to alternate between generating text and video frames, allowing the model to ``think in words'' about subsequent content before ``acting in pixels'' to produce frames. 
This design offloads much of the responsibility for deciding what should happen next to the language modeling tower, enabling improved visual quality and prompt alignment of generated videos. It also enables fine-grained controllability, allowing users to modify the video generation trajectory through text interventions at any point in the process.
In controlled experiments on video game data, \name demonstrates substantial improvements in both visual quality (preferred $91\%$ of the time in human evaluations vs. a comparable text-to-video model) and controllability (19 point improvement % 0.782 vs 0.591
in fine-grained instruction following accuracy vs. a ``think-then-act'' approach). \name also scales to natural videos, as we show by augmenting sports videos with interleaved natural language action descriptions using vision-language models (VLMs). 
Training \name on this corpus yields strong visual quality and prompt alignment, showcasing the model's ability to reason about and generate complex real-world action sequences.
Together, these results highlight \name as a promising step toward video generation with open-ended textual reasoning and control. 
}

%% file: sections/01_intro.tex
\section{Introduction}\label{sec:intro}

\begin{figure}[t]
\centering
\begin{tikzpicture}[
    >=stealth,  % Set default arrow tip style
    textbox/.style={rectangle, draw=yellow!80, fill=yellow!20, thick,
                    minimum width=11cm, minimum height=1cm, text width=10.7cm, align=justify, font=\small, inner sep=0.15cm},
    bluetextbox/.style={rectangle, draw=orange!80!red!60, fill=orange!20!red!10, thick,
                    minimum width=11cm, minimum height=1cm, text width=10.7cm, align=left, font=\small, inner sep=0.15cm},
    imagebox/.style={rectangle, draw=gray!60, thick, inner sep=0pt},
    timestamp/.style={circle, draw=gray!60, fill=gray!20, thick, minimum size=0.65cm}
]
% Row 0 - Blue Text (Title/Header)
\node[bluetextbox] (text0) at (2.5,2) {\textbf{Initial user prompt:} A surfer in a red top and black shorts skillfully rides a large wave.};
\node[font=\Large, text=orange!80!red!60] at (-3.8,2) {\faUser};
% Dotted line separator with gap for text
\draw[dotted, thick, gray!60] (-2.5,0.65) -- (0.0,0.65);
\node[font=\small, text=black!70] at (3.5,0.65) {\emph{Interleaved text and video generation process}};
\draw[dotted, thick, gray!60] (7.0,0.65) -- (9.5,0.65);
% Row 1 - Text
\node[textbox] (text1) at (3.5,-0.5) {\textbf{Model generated:} The surfer rides steadily along the wave's face, extending their arms for stability.};
\node[font=\Large, text=yellow!80!black!60] at (10.0,-0.5) {\faRobot};
% Row 2 - Image Frames
\node[imagebox] (img1) at (0.5,-2) {\includegraphics[width=2.5cm, height=1.5cm]{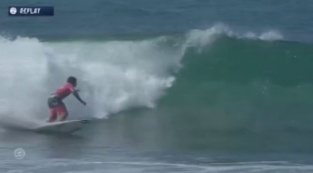}};
\node at (2.25,-2) {$\cdots$};
\node[imagebox] (img2) at (4,-2) {\includegraphics[width=2.5cm, height=1.5cm]{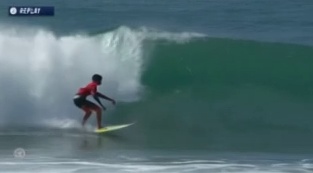}};
\node at (5.75,-2) {$\cdots$};
\node[imagebox] (img3) at (7.5,-2) {\includegraphics[width=2.5cm, height=1.5cm]{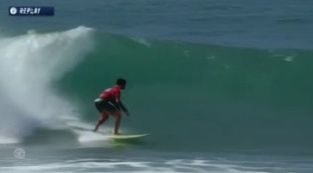}};
\node[font=\Large, text=yellow!80!black!60] at (10.0,-2) {\faRobot};
% Row 3 - Text
% \node[bluetextbox] (text2) at (2.5,-3.5) {\textbf{User intervention:} The surfer leans into a turn, carving along the wave's face with precision. The board glides smoothly, and the surfer extends an arm for balance as the wave curls and sprays water.};
\node[bluetextbox] (text2) at (2.5,-3.5) {\textbf{User intervention:} The surfer leans into an upward turn.};
\node[font=\Large, text=orange!80!red!60] at (-3.8,-3.5) {\faUser};
% Row 4 - Image Frames
\node[imagebox] (img4) at (2.25,-5) {\includegraphics[width=2.5cm, height=1.5cm]{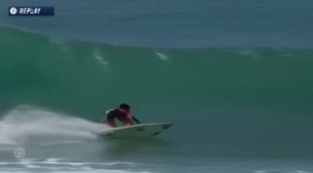}};
\node at (4,-5) {$\cdots$};
\node[imagebox] (img5) at (5.75,-5) {\includegraphics[width=2.5cm, height=1.5cm]{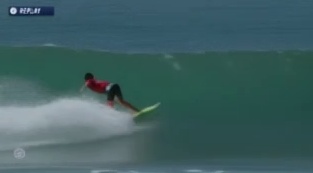}};
\node[font=\Large, text=yellow!80!black!60] at (10.0,-5) {\faRobot};
% Row 5 - Text
% \node[textbox] (text3) at (4.5,-6.5) {\textbf{Model generated:}  The surfer performs a sharp turn, lifting the board slightly as they ride up the wave's face. The surfer's body leans into the motion, maintaining control as the wave crashes behind.};
\node[textbox] (text3) at (3.5,-6.5) {\textbf{Model generated:}  The surfer performs a sharp downward turn, lifting the board above the wave. The waves crash behind them.};
\node[font=\Large, text=yellow!80!black!60] at (10.0,-6.5) {\faRobot};
% Row 6 - Image Frames
\node[imagebox] (img7) at (0.5,-8) {\includegraphics[width=2.5cm, height=1.5cm]{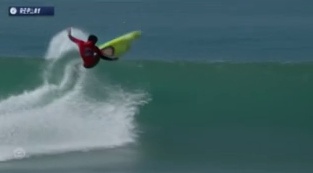}};
\node at (2.25,-8) {$\cdots$};
\node[imagebox] (img8) at (4,-8) {\includegraphics[width=2.5cm, height=1.5cm]{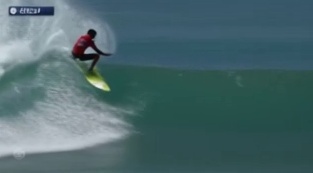}};
\node at (5.75,-8) {$\cdots$};
\node[imagebox] (img9) at (7.5,-8) {\includegraphics[width=2.5cm, height=1.5cm]{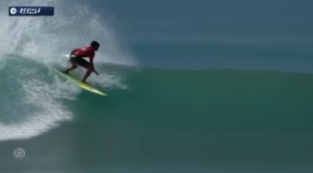}};
\node[font=\Large, text=yellow!80!black!60] at (10.0,-8) {\faRobot};
% Timeline arrow on the right (arrows on both ends)
\draw[->, gray!70, line width=2pt] (11.0,2.5) -- (11.0,-9);
% Tick marks on arrow
\draw[gray!70, line width=1pt] (10.85,-0.5) -- (11.15,-0.5);
\draw[gray!70, line width=1pt] (10.85,-3.5) -- (11.15,-3.5);
\draw[gray!70, line width=1pt] (10.85,-6.5) -- (11.15,-6.5);
% Timestamps (only by image frame rows)
\node[font=\footnotesize, text=gray!70] at (11.8,-0.5) {\emph{t=0.0s}};
\node[font=\footnotesize, text=gray!70] at (11.8,-3.5) {\emph{t=1.6s}};
\node[font=\footnotesize, text=gray!70] at (11.8,-6.5) {\emph{t=2.4s}};
\end{tikzpicture}
    \caption{\textbf{Overview of \name interleaved text and video generation.} \name is a unified generative modeling framework which decomposes video generation into an interleaved text and video generation process.  During inference, \name dynamically alternates between autoregressively generating plans in text and semi-autoregressively generating chunks of video frames, allowing the model to \emph{think in words} about the content of the subsequent frames before \emph{acting in pixels} to produce those frames. This framework enables fine-grained and flexible control during video generation, allowing users to potentially intervene and modify the video generation trajectory at any point through textual prompting.}
\label{fig:teaser_timeline}
\end{figure}
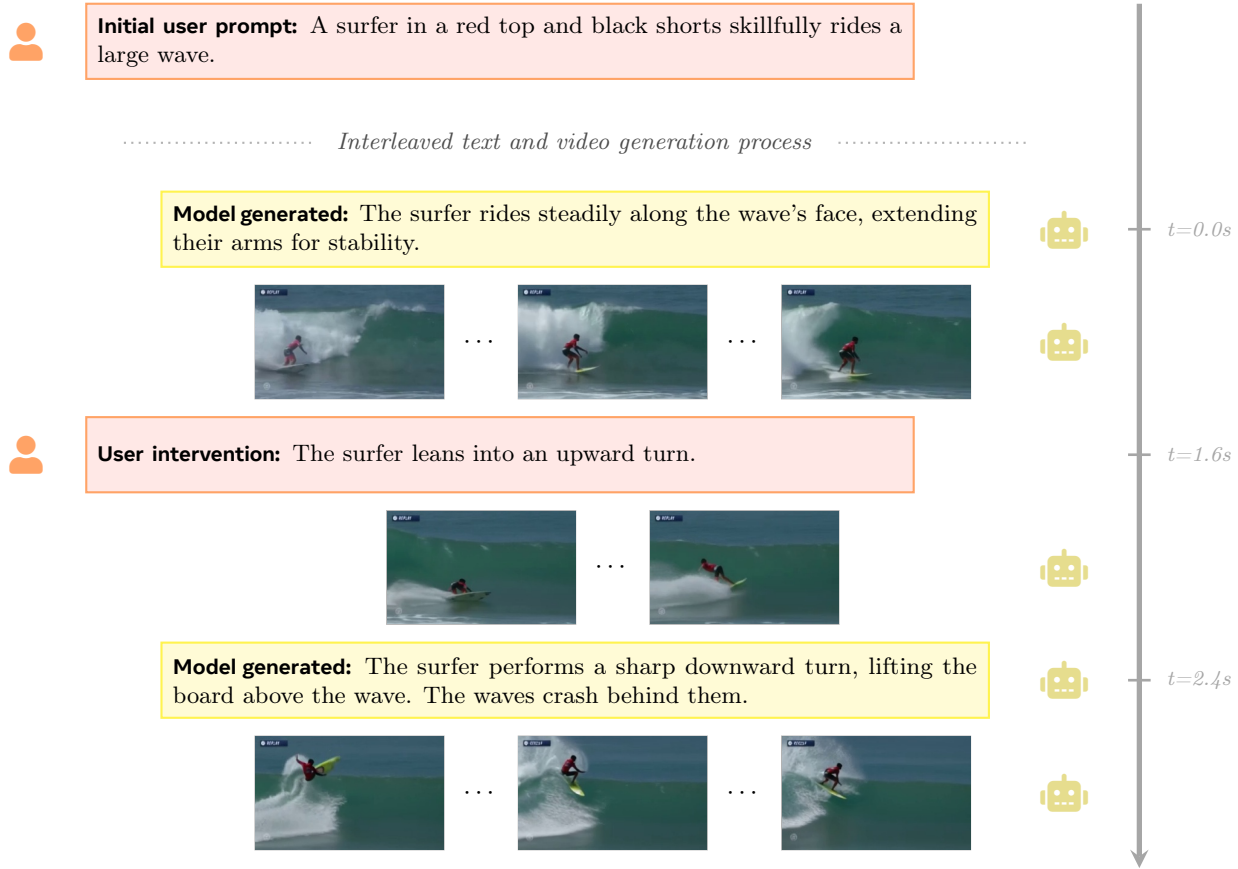

Despite incredible progress in visual quality, video generation models can still struggle with complex outputs that
require significant semantic branching or repeated high-level reasoning about what should happen next. In this paper, we introduce a new class of omni video-text models that integrate ideas from recent LM reasoning advances to address this challenge. Our approach generalizes previous omni models that have focused on text and image modalities~\citep{chameleonteam2024chameleonmixedmodalearlyfusionfoundation,transfusion,wu2024janusdecouplingvisualencoding,deng2025bagel,li2025manzano} as well as interactive video generations models like Genie \citep{bruce2024genie} that require explicit user input at each step. We instead show that it is possible to train an omni model which automatically decomposes video generation into an interleaved text and video generation process, thereby significantly improving quality and controllability. 

We present \name, which is a Transfusion-style modeling approach \citep{transfusion} that jointly learns language modeling (next-token prediction) and video flow matching \citep{liu2022flow,esser2024scaling} (next-frame prediction). 
At inference time, \name dynamically alternates between generating text and generating chunks of video frames, allowing the model to \emph{think in words} about the content of the subsequent frames before \emph{acting in pixels} to produce those frames. This approach offloads much of the semantic decision-making to the language modeling component of the model, capitalizing on recent advances in LLM reasoning capabilities and reducing the entropy of the video generation process.
The design also affords strong controllability: users can inspect, edit, or steer the textual plan to modify the video generation trajectory at any timestep.

\name adopts a Mixture-of-Transformers (MoT) \citep{liang2024mixture} architecture with dedicated towers for video and text modalities, enabling modality-specific processing while maintaining a global self-attention over the entire multimodal input sequence. Following LMFusion \citep{lmfusion}, we initialize the text tower from a pre-trained language model.
\name is trained on interleaved sequences of text and temporally segmented chunks of video frames. We employ bi-directional attention within video frame chunks and causal attention otherwise.

We evaluate performance in two domains. First, we use video game data from \emph{Counter Strike; Global Offensive} (\CSGO) curated and open-sourced by \citet{pearce2022csgo}, which provides strongly correlated interleaved data via controller actions (represented as text) and resulting video gameplay. 
Controlled experiments on this data show that \name substantially outperforms competing approaches in both visual quality -- preferred $91\%$ of the time over a comparable \TtoV baseline -- and controllability, with a 19 point improvement in fine-grained instruction-following accuracy compared to a ``think-then-act'' (\ThinktoV) method which generates a detailed text action plan prior to generating video. 
Relatedly, action-conditioned world models \citep[e.g.,][]{bruce2024genie} can also generate game or synthetic video-world data. However, having them automatically generate videos without dense human control requires either a separate controller model \citep[e.g.,][]{raad2024scaling} or a costly planning algorithm \citep[e.g.,][]{bar2025navigation}. Our \name approach performs action generation and video-world generation flexibly and end-to-end.  

Next, we demonstrate how to scale this modeling paradigm to real-world videos which typically lack such temporally-aligned text data, by augmenting video data with interleaved natural language action descriptions using vision-language models (VLMs). Using such a pipeline, we curate 8K hours of interleaved text and video content in the sports domain, chosen for its dynamic motion and rich action content, and demonstrate that training \name on this corpus yields strong visual quality and prompt alignment relative to both established external video generation models and controlled \TtoV and \ThinktoV baselines. In holistic preference evaluations, \name is favored 54.0\% of the time compared to \TtoV (34.7\% unfavorable) and 53.3\% of the time compared to \ThinktoV (41.3\% unfavorable). We also present qualitative examples showcasing how users can dynamically steer video generation through intermediate textual prompting.

Together, these experiments highlight \name as a promising step towards unifying advances in language model reasoning with highly controllable video generation systems, leveraging natural language not merely as input conditioning, but as an active reasoning mechanism for decomposing complex visual and temporal tasks.

Our contributions are summarized as follows:
\begin{itemize}
    \item \textbf{\name} (\autoref{sec:methods}): We introduce \name, a unified generative modeling framework capable of decomposing video generation into an interleaved text and video generation process. 
    \item \textbf{Controlled Experiments with Video Game Data} (\autoref{sec:findings}) We validate this approach through controlled experiments with video game (\CSGO) data, demonstrating that \name outperforms \TtoV and \ThinktoV baselines in both visual quality and controllability. Specifically, in pairwise human evaluations, videos generated by \name are preferred to those generated by a \TtoV baseline 91\% of the time, and \name shows a 19 point improvement in fine-grained instruction following accuracy compared to a \ThinktoV baseline.
    \item \textbf{Scaling \name to Real World Data} (\autoref{sec:soccer}) Finally, we scale this paradigm to real world video data by synthetically augmenting sport video data with interleaved captions. Training \name on this corpus yields strong prompt alignment and visual quality relative to both external and controlled baselines (wins 54.0\% vs. 34.7\% and 53.3\% vs. 41.3\% against comparable \TtoV and \ThinktoV baselines, respectively), showcasing \name's ability to seamlessly reason about and generate complex visual action sequences. 
\end{itemize}

%% file: sections/03_methods.tex
\section{\name}\label{sec:methods}

\begin{figure}[ht]
    \centering
    \includegraphics[width=1.0\linewidth]{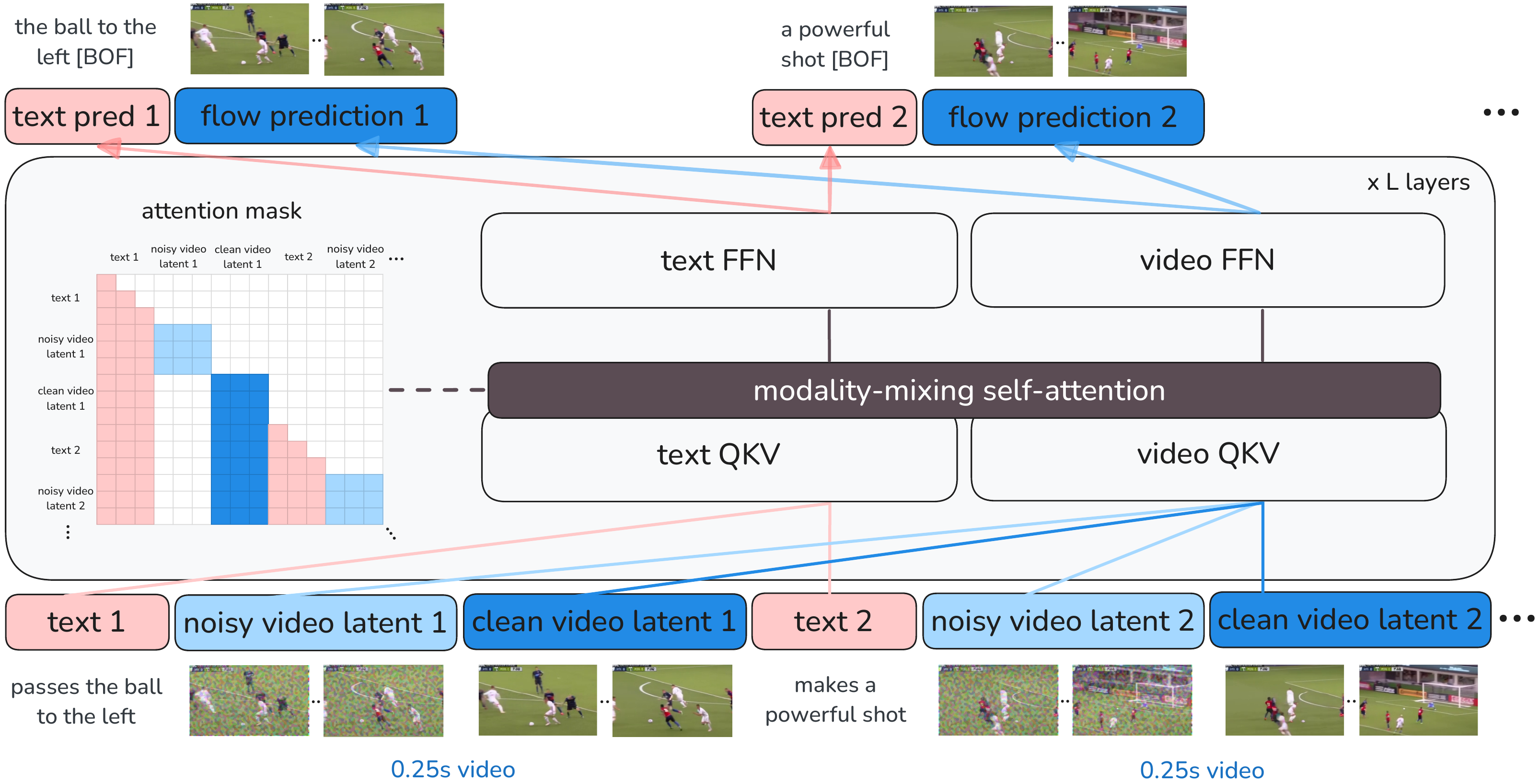}
    \caption{\textbf{Overview of \name architecture.} \name builds on the Transfusion \citep{transfusion} modeling approach, jointly learning language modeling and video flow matching during training. \name autoregressively generates interleaved chunks of video frames and text tokens, maintaining strict temporal causality: each token or chunk of frames can only attend to preceding tokens or chunks. \name adopts a Mixture-of-Transformers (MoT) \citep{liang2024mixture} architecture with dedicated towers for the video and text modalities.
    }
    \label{fig:teaser}
    \vspace{-0.1em}
\end{figure}

We present \name, a unified modeling framework which decomposes video generation into an interleaved text and video generation process. 

\name builds on the Transfusion-style \citep{transfusion} approach, jointly learning language modeling (next-token prediction) and video flow matching (next-frame prediction). 
At a global level, \name autoregressively generates interleaved chunks of video frames and text tokens, maintaining strict temporal causality: each token or chunk of frames conditions only on preceding tokens or chunks. At the local level, within each chunk comprising one or several video frames, the model is non-autoregressive and uses a flow matching objective. This autoregressive video generation component, when considered independently of the interleaved text, follows a design similar to MAGI-1 \citep{teng2025magi}.

\name adopts a Mixture-of-Transformers (MoT) \citep{liang2024mixture} architecture with dedicated towers for the video and text modalities, enabling modality-specific processing while maintaining a global self-attention over the entire multimodal input sequence. Following \citet{lmfusion}, we initialize the text tower with a pretrained language model. See \autoref{fig:teaser} for an overview of TV2TV’s architecture during training. At inference time, \name flexibly alternates between text and video generation, factorizing the video generation into interleaved textual planning and video segment generation.  

In the following section, we detail \name's interleaved data representation (\autoref{subsec:method_data_representation}), architecture and optimization (\autoref{subsec:method_architecture}), inference procedure (\autoref{sec:method_inference}), and task formulation (\autoref{subsec:method_task}).

\subsection{Data representation}\label{subsec:method_data_representation}

\name models both discrete text and continuous videos in an interleaved fashion. Below, we describe how we prepare these interleaved sequences.

\paragraph{Discrete text tokens.}
To represent language, we use a regular BPE tokenizer to obtain discrete text tokens.\footnote{Specifically, we use the \texttt{tiktoken}-based tokenizer from \citet{llama3}.} 

\paragraph{Continuous video tokens.} To represent video, we use a VAE tokenizer with a causal 3D CNN backbone to obtain continuous video tokens.\footnote{Specifically, we use the \texttt{Cosmos-Tokenize1-CV4x8x8-360p} tokenizer from \citet{agarwal2025cosmos}.} 
Our video tokenizer has a temporal compression factor of 4, so every 4 frames are grouped together as an atomic \emph{chunk} of frames, except for the first frame. 
The video tokenizer was trained with sequences of 49 frames, so we chunk and pad longer videos to multiple 49 frames, pass them through the tokenizer, and obtain a series of \emph{latent frames} in the latent space, one for each atomic chunk of frames. 
We work with 16 FPS videos throughout this work, so each latent frame corresponds to 0.25 seconds of video. 
For simplicity, we use \emph{frame chunk} to refer to such 0.25-second video in the latent space from now.

\paragraph{Interleaved text and video sequences.}
In the interleaved sequence, we organize text segments and frame chunks chronologically according to their timestamps. Video frame chunks are timestamped by their start time, while text segments are assigned timestamps as defined in \autoref{sec:findings} and \autoref{sec:soccer}.
Our generation process respects temporal causality: each text segment or video frame chunk is conditioned only on content from earlier timestamps. When a text segment and video frame chunk share the same timestamp, we place the text before the frame chunk in the sequence such that the video generation can condition on the associated text (i.e., allowing the model to \emph{think} in text before \emph{acting} to generate those frames). Additionally, we introduce two special discrete tokens -- \emph{beginning-of-frame} (BOF) and \emph{end-of-frame} (EOF) -- that delimit the continuous tokens of each video frame chunk. This design enables the model to automatically transition between text generation mode and video generation mode during inference (see \autoref{sec:method_inference} for details).

\paragraph{Clean and noisy latents.}
Though across the interleaved sequence we predict text and video frame chunks autoregressively, within each frame chunk we perform flow matching on the continuous video tokens. 
Flow matching and diffusion methods require noisy, interpolated input representations for training. However, autoregressive generation under teacher-forcing requires access to clean representations from previous sequence elements to maintain proper conditioning context.
To resolve this conflict, we maintain two copies of each video frame's representation in the input sequence: a noisy frame chunk followed immediately by a clean frame chunk.
This design allows the model to condition on clean historical context while learning to denoise current frames. This contrasts with MAGI-1 \citep{teng2025magi}, which addresses this challenge by enforcing monotonicity in noise, ensuring earlier video chunks are cleaner than later ones. While this maintains a single representation per frame, it does not work well in our interleaved text-video setting as it limits the model's ability to effectively interact with and condition on the textual components of the sequence.

\paragraph{Notation.}
In summary, the text and video data are tokenized and prepared as an interleaved sequence of discrete and continuous tokens. We refer to the text, noisy video frame chunk, and clean video frame chunk representations as $\boldsymbol{x}^{\text{txt}}$, $\boldsymbol{x}^{\text{noisy-vid}}$, and $\boldsymbol{x}^{\text{clean-vid}}$, respectively. 
Thus, each sequence for our \name training is in the form: 
\begin{align}
    [\boldsymbol{x}_1^{\text{txt}}, \boldsymbol{x}_1^{\text{noisy-vid}}, \boldsymbol{x}_1^{\text{clean-vid}}, \boldsymbol{x}_2^{\text{txt}}, \boldsymbol{x}_2^{\text{noisy-vid}}, \boldsymbol{x}_2^{\text{clean-vid}}, \ldots , \boldsymbol{x}_N^{\text{txt}}, \boldsymbol{x}_N^{\text{noisy-vid}}, \boldsymbol{x}_N^{\text{clean-vid}}]
\end{align}
where $N$ is the total number of latent video frames. 
As a shorthand for indexing specific modalities, we use $\boldsymbol{x}^{\text{txt}} = \oplus_{i=1}^{N} \boldsymbol{x}_{i}^{\text{txt}}$ for the text tokens in the sequence, $\boldsymbol{x}^{\text{vid}} = \oplus_{i=1}^{N} (\boldsymbol{x}_{i}^{\text{noisy-vid}}, \boldsymbol{x}_{i}^{\text{clean-vid}})$ for the video tokens in the sequence, and $\boldsymbol{x}^{\text{all}} = \oplus_{i=1}^{N} (\boldsymbol{x}_{i}^{\text{txt}}, \boldsymbol{x}_{i}^{\text{noisy-vid}}, \boldsymbol{x}_{i}^{\text{clean-vid}})$ for the full interleaving sequence. $\oplus$ indicates concatenation along the sequence dimension.

\subsection{Architecture and optimization}\label{subsec:method_architecture}
\name adopts an MoT \citep{liang2024mixture} architecture with dedicated transformer towers for each video and text. Following \citet{lmfusion}, the text tower is initialized from a pretrained LM. During training, \name jointly learns language modeling (next-token prediction) and video flow matching (next-frame prediction).  We provide further architectural and optimization details below. 

\paragraph{Noise interpolation.}
Flow matching on our continuous video frame representations requires interpolating them with pure noise samples to produce the noisy latents. Specifically, we use rectified flow \citep{liu2022flow} in a configuration similar to \citet{esser2024scaling}. Let $\boldsymbol{x}^{\text{clean-vid}}$ be the video latents encoded by the continuous video tokenizer. 
\begin{align}
    \boldsymbol{x}^{\text{noisy-vid}} = t\boldsymbol{x}^{\text{clean-vid}} + (1-t) \boldsymbol{\epsilon}
\end{align}
where $t$ is sampled from a logit-normal distribution, $t \sim \operatorname{logistic}(\mathcal{N}(\mu, \sigma^2))$,\footnote{We use $\mu=0$ and $\sigma=1.4$ in this work.} and $\boldsymbol{\epsilon}$ is a Gaussian noise, $\boldsymbol{\epsilon} \sim \mathcal{N}(\mathbf{0}, \mathbf{I})$.

\paragraph{Dropout.}
To enable text classifier-free guidance (CFG) during inference, we drop out $\boldsymbol{x}_{i}^{\text{txt}}$ randomly with a small probability $p_{\text{txt-drop}}$. 
To alleviate exposure bias in the teacher-forcing sequential training, a popular strategy is to inject a small amount of noise into clean data \citep{teng2025magi}.
In our case, we apply a soft dropout for $\boldsymbol{x}^{\text{clean-vid}}$, flipping $\boldsymbol{x}_{i}^{\text{clean-vid}}$ to $\boldsymbol{x}_{i}^{\text{noisy-vid}}$ with a small probability $p_{\text{clean-vid-flip}}$.

\paragraph{Input projection.}
The input text tokens $\boldsymbol{x}^\text{txt}$ are projected by a linear embedding layer to a sequence of text hidden states $\boldsymbol{h}_\text{in}^\text{txt}$. 
The video frame latents $\boldsymbol{x}^{\text{vid}}$ are projected to a sequence of video hidden states $\boldsymbol{h}_\text{in}^\textit{vid}$ via a U-Net downsampler \citep{ronneberger2015u}. 
The timestep $t$ is integrated in the downsampler via adding a time embedding. For each token in the video latents $\boldsymbol{x}^{\text{vid}}$ corresponding to different spatial patches, we also add an absolute 2D position embedding through the downsampler. 
\begin{align}
    \boldsymbol{h}_{\text{in}}^{\text{txt}} &= \operatorname{Proj}_{\text{txt}}(\boldsymbol{x}^{\text{txt}}) \\ 
    \boldsymbol{h}_{\text{in}}^{\text{vid}} &= \operatorname{UNet\text{-}Down}_{\text{vid}}(\boldsymbol{x}^{\text{vid}}, t)
\end{align}

\paragraph{Modality-specific self-attention.}
Following the original MoT design \citep{liang2024mixture}, the text hidden states $\boldsymbol{h}_\text{in}^\text{txt}$ and video hidden states $\boldsymbol{h}_\text{in}^\text{vid}$ are transformed to their respective queries, keys, and values via separate $Q, K, V$ matrices. The pre-attention layer normalization is also modality-specific and is folded into the $\operatorname{QKV}$ functions in the equations below for simplicity. 
\begin{align}
    \boldsymbol{h}_{\text{Q}}^{\text{txt}} &= \operatorname{Q}_{\text{txt}}(\boldsymbol{h}_{\text{in}}^{\text{txt}}), &
    \boldsymbol{h}_{\text{K}}^{\text{txt}} &= \operatorname{K}_{\text{txt}}(\boldsymbol{h}_{\text{in}}^{\text{txt}}), &
    \boldsymbol{h}_{\text{V}}^{\text{txt}} &= \operatorname{V}_{\text{txt}}(\boldsymbol{h}_{\text{in}}^{\text{txt}}) \\
    \boldsymbol{h}_{\text{Q}}^{\text{vid}} &= \operatorname{Q}_{\text{vid}}(\boldsymbol{h}_{\text{in}}^{\text{vid}}), &
    \boldsymbol{h}_{\text{K}}^{\text{vid}} &= \operatorname{K}_{\text{vid}}(\boldsymbol{h}_{\text{in}}^{\text{vid}}), &
    \boldsymbol{h}_{\text{V}}^{\text{vid}} &= \operatorname{V}_{\text{vid}}(\boldsymbol{h}_{\text{in}}^{\text{vid}})
\end{align} 
Attention is then computed across all tokens in the interleaving sequence. The attention-weighted values are projected back to the hidden state dimension using modality-specific $\operatorname{O}$ matrices. 
\begin{align}
    \boldsymbol{h}_{\text{O}}^{\text{txt}} &= \operatorname{O}_{\text{txt}}\!\left(
        \operatorname{softmax}\!\left(
            \frac{\operatorname{mask}(\boldsymbol{h}_{\text{Q}}^{\text{txt}}\,{\boldsymbol{h}_{\text{K}}^{\text{all}}}^{\;\!T})}{\sqrt{d}}
        \right)
        \boldsymbol{h}_{\text{V}}^{\text{all}}
    \right) \\
    \boldsymbol{h}_{\text{O}}^{\text{vid}} &= \operatorname{O}_{\text{vid}}\!\left(
        \operatorname{softmax}\!\left(
            \frac{\operatorname{mask}(\boldsymbol{h}_{\text{Q}}^{\text{vid}}\,{\boldsymbol{h}_{\text{K}}^{\text{all}}}^{\;\!T})}{\sqrt{d}}
        \right)
        \boldsymbol{h}_{\text{V}}^{\text{all}}
    \right)
\end{align} 
where $\operatorname{mask}$ denotes a hybrid attention mask---applying a causal mask to the positions of text tokens and a block-causal mask to the positions of noisy and clean video tokens. 
An additional principle for masking is that noisy video tokens cannot be attended by any future tokens in the sequence. 
A global 1D RoPE is also applied here to all positions for all modalities, with duplicated position IDs for noisy and clean video tokens of the same frame chunk.

\paragraph{Modality-specific feed-forward network.}
Again, following the original MoT \citep{liang2024mixture} design, after self-attention, we use modality-specific FFNs to further transform text and video representations separately. The pre-FFN layer normalization is also modality-specific and is folded in the $\operatorname{FFN}$ function for simplicity.\footnote{We also do not show residual connections for simplicity of notation.} 
\begin{align}
    \boldsymbol{h}_{\text{FFN}}^{\text{txt}} &= \operatorname{FFN}_{\text{txt}}(\boldsymbol{h}_{\text{O}}^{\text{txt}}) \\
    \boldsymbol{h}_{\text{FFN}}^{\text{vid}} &= \operatorname{FFN}_{\text{vid}}(\boldsymbol{h}_{\text{O}}^{\text{vid}})
\end{align}

\paragraph{Output projection.}
After $L$ layers of self-attention and FFNs, the resulting hidden states are projected either to logits in text via an output embedding or to a predicted flow via a U-Net upsampler.
\begin{align}
    \boldsymbol{s}_{\text{logits}}^{\text{txt}} &= \operatorname{LM\text{-}Head}_{\text{txt}}(\boldsymbol{h}_{\text{FFN}}^{\text{txt}}) \\
    \boldsymbol{v}_{\text{pred}}^{\text{noisy-vid}} &= \operatorname{UNet\text{-}Up}_{\text{vid}}(\boldsymbol{h}_{\text{FFN}}^{\text{noisy-vid}}, \boldsymbol{h}_{\text{in}}^{\text{noisy-vid}}, t)
\end{align}

\paragraph{Training objective.}
The training objective is a combination of text cross entropy loss and video flow loss. All model parameters are optimized jointly using both losses.  
\begin{align}
    \mathcal{L}_{\text{txt}}
    &= \operatorname{CE}(\boldsymbol{s}_{\text{logits}}^{\text{txt}},{\boldsymbol{x}^{\text{txt}}}) \\
    \mathcal{L}_{\text{vid}}
    &= \operatorname{MSE}\bigr(\boldsymbol{v}_{\text{pred}}^{\text{noisy-vid}}
    , (\boldsymbol{x}^{\text{clean-vid}} - \boldsymbol{\epsilon}) \bigr) \\
    \mathcal{L}
    &= \lambda_{\text{txt}} \mathcal{L}_{\text{txt}} + \lambda_{\text{vid}} \mathcal{L}_{\text{vid}}
\end{align}

\paragraph{Parameter initialization.} 
Due to the modality-specific design above, we have a separate set of parameters for each modality. The text tower's parameters $\theta_{\text{txt}}$ include the parameters from $\operatorname{Proj}_{\text{txt}}$, $\operatorname{Q}_{\text{txt}}$, $\operatorname{K}_{\text{txt}}$, $\operatorname{V}_{\text{txt}}$, $\operatorname{O}_{\text{txt}}$, $\operatorname{FFN}_{\text{txt}}$, and $\operatorname{LM\text{-}Head}_{\text{txt}}$. The video tower's parameters $\theta_{\text{vid}}$ include the parameters from $\operatorname{UNet\text{-}Down}$, $\operatorname{Q}_{\text{vid}}$, $\operatorname{K}_{\text{vid}}$, $\operatorname{V}_{\text{vid}}$, $\operatorname{O}_{\text{vid}}$, $\operatorname{FFN}_{\text{vid}}$, and $\operatorname{UNet\text{-}Up}_{\text{vid}}$. 
Following \citet{lmfusion}, we initialize $\theta_{\text{txt}}$ from a pre-trained Llama model.

\subsection{Inference}\label{sec:method_inference}

\name's key innovation lies in its ability to dynamically switch between text and video generation during inference.
This is achieved by using a special \emph{beginning-of-frame} (BOF) token which controls the transition from text to video generation. 
By default, \name operates in text mode, generating tokens autoregressively like standard LLMs. At each autoregressive step $i-1$, the model samples a next token $x_{i}^{\text{txt}}$ and proceeds as follows:
\begin{itemize}

    \item \textbf{Text token  ($x_{i}^{\text{txt}} \in V$):} If the current token $x_{i}^{\text{txt}}$ is within the regular language vocabulary $V$ (not special token BOF or EOS), the model continues autoregressive text generation. 
        
    \item \textbf{BOF token ($x_{i}^{\text{txt}} = \texttt{BOF}$):} If the current token is the BOF token, the model 
        \begin{enumerate}
            \item Extends the sequence with tokens representing one video frame chunk ($\boldsymbol{x}_{i}^{\text{noisy-vid}}$). 
            \item Initializes these tokens from a normal distribution $\mathcal{N}(\mathbf{0}, \mathbf{I})$.
            \item Runs $m$ steps of an ODE solver (e.g., Euler) using $\boldsymbol{x}_{i}^{\text{noisy-vid}}$ and the KV cache of the previous tokens (optionally applying CFG).\footnote{When CFG is enabled during inference, we maintain both a text-conditional sequence $[\boldsymbol{x}_1^{\text{txt}}, \boldsymbol{x}_1^{\text{clean-vid}}, \ldots , \boldsymbol{x}_{i-1}^{\text{txt}}, \boldsymbol{x}_{i-1}^{\text{clean-vid}}, \boldsymbol{x}_i^{\text{txt}}, \boldsymbol{x}_i^{\text{noisy-vid}}]$ and a text-unconditional sequence $[\boldsymbol{x}_1^{\text{clean-vid}}, \ldots , \boldsymbol{x}_{i-1}^{\text{clean-vid}}, \boldsymbol{x}_i^{\text{noisy-vid}}]$, and take ODE steps contrastively on the two $\boldsymbol{x}_i^{\text{noisy-vid}}$.} 
            \item Run forward pass with the final output of the solver $\boldsymbol{x}_{i}^{\text{clean-vid}}$ and updates the KV cache, and then resumes autoregressive text generation.
        \end{enumerate}

    \item \textbf{End of sequence ($x_{i}^{\text{txt}} = \texttt{EOS}$ or context length exceeded):} The generation process terminates.

\end{itemize}

\paragraph{Extending sequences beyond the trained context length} Because \name is globally autoregressive,  it can natively generate videos longer than its trained context length using sliding windows. 
To extend an interleaved text-video sequence $\oplus_{i=1}^{N} (\boldsymbol{x}_{i}^{\text{txt}}, \boldsymbol{x}_{i}^{\text{vid}})$, we retain the second half $\oplus_{i=N/2}^{N} (\boldsymbol{x}_{i}^{\text{txt}}, \boldsymbol{x}_{i}^{\text{vid}})$ and use it as a condition $\oplus_{i=1}^{N/2} (\boldsymbol{x}_{i}^{\text{txt}}, \boldsymbol{x}_{i}^{\text{vid}})$ for the next generation window. 

% \medskip

\subsection{Task formulation}\label{subsec:method_task}
While thus far we have presented a general approach for training on and generating interleaved text and video sequences, in this paper, we focus on \emph{video generation} as the primary objective while leveraging text generation as an auxiliary task that provides semantic guidance. 
Rather than generating complex video content directly, the \name framework decomposes the video generation process into interleaved text and video sequences. This factorization offers two advantages. First, it offloads  much of the semantic complexity to the text generation components of the model, reducing the burden on the video generation component.
Second, it enables flexible user control during generation,  allowing users to intervene and modify the video generation trajectory at any point through textual prompting. 

A central question is how to obtain such interleaved text and video data for training. Video game data with associated controller actions provides a natural source, as the controller action can act as the textual ``plan'' for the subsequent video frames. \autoref{sec:findings} presents controlled experiments with such data. 

To extend this modeling paradigm to general domain video data, which largely lacks clean and temporally-aligned captions, in \autoref{sec:soccer}, we experiment with a methodology for synthetically augmenting real-world videos with interleaved natural language action descriptions using vision-language models (VLMs).

%% file: sections/04_findings.tex
\section{Controlled Experiments with Video Game Data}\label{sec:findings}

In this section, we evaluate \name using video gameplay footage (video) paired with controller actions (text). Video games offer an ideal testbed for interleaved text and video approaches: controller inputs serve as textual ``plans'' that reflect the subsequent actions displayed in the gameplay footage. By using the same video data while varying the text representations, we can directly evaluate whether letting the model ``think in text'' \emph{interleavingly} before ``acting in pixels'' improves video generation quality. Additionally, we investigate whether this approach enables fine-grained user control during inference.

Specifically, we design our experiments to address the following research questions: 
\begin{itemize}
\item \textbf{Overall visual quality:} Does generating interleaved planning text improve overall video quality compared to non-interleaved baselines? 
\item \textbf{Fine-grained controllability:} Does training with interleaved text improve the user controllability of video generation, i.e., can a mid-sequence text intervention reliably steer the video? 
\end{itemize} 
We train \name and two baselines on gameplay video, actions, and metadata from the \emph{Counter-Strike: Global Offensive} (\CSGO) dataset curated and open-sourced by \citet{pearce2022csgo}. 

\subsection{Modeling details}\label{subsec:thinktov_ttov}

We compare \name with two controlled baselines, which we refer to as \TtoV and \ThinktoV, to isolate different aspects of our approach:\footnote{During inference, all models receive both the meta-prompt and a starting frame. We use the shorthand \TtoV and \ThinktoV (rather than TI2V and ThinkI2V) for simplicity.} 
\begin{itemize}
    \item \textbf{\TtoV:} We adopt the same autoregressive video modeling framework as \name, but the model is trained without any interleaved text, conditioning solely on the meta-prompt (and an initial starting frame).
    \item \textbf{\ThinktoV:} Rather than generating text and video in an interleaved fashion, after receiving the meta-prompt, \ThinktoV first generates a detailed roll-out of all subsequent text actions \emph{before} producing any video frames. In other words, text generation (`\emph{thinking}') is followed by video generation in a sequential, non-interleaved manner. Again, we adopt the same autoregressive video modeling framework as \name.  
\end{itemize}
See \autoref{fig:csgo_data} for an illustration of the different sequence representations used for \name, \TtoV, and \ThinktoV. 

For these experiments, all models adopt a 3B-MoT backbone with modality-specific 3B-parameter text and video towers. The text tower is initialized with Llama-3.2-3B \citep{llama3}. We train for 50K steps with a batch size of 128, utilizing a cosine learning rate scheduler with a maximum learning rate of  $3 \mathrm{e}{-4}$. Detailed model and training configurations are provided in \autoref{tab:model_configs} in \autoref{appendix:configuration_csgo}.

\subsection{Data details}

\begin{figure}
  \centering
    \includegraphics[width=0.80\columnwidth]{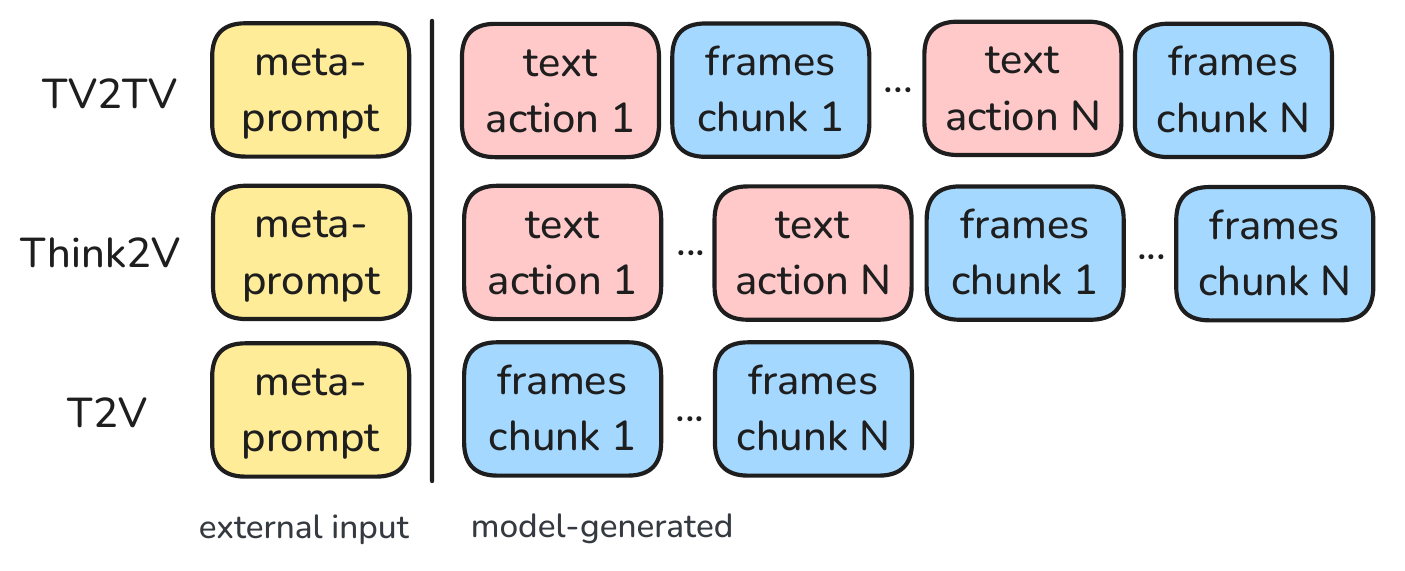}
    \caption{\textbf{Illustration of sequence representations for \name, \TtoV, and \ThinktoV model variants}. 
    During inference, the meta-prompt (and a single conditioning frame) is provided and the rest of the sequence is model-generated.}
\label{fig:csgo_data}
\end{figure}

All models are trained on 95 hours of video from the video game \emph{Counter-Strike; Global Offensive} (\CSGO) curated and open-sourced by \cite{pearce2022csgo}. Each video frame is associated with a controller action, but as we group each 4 frames into a single video latent, we concatenate 4 controller actions, stringify them as text, and pass them to the model, e.g.: 
\begin{center}
\begin{minipage}{0.25\textwidth}
\centering
\emph{(w, d, shift). 10, 0. 0, 0.\\
(d). 4, 0. 0, 0.\\
(d). 4, 0. 0, 0.\\
(d). 0, 0. 1, 0.}
\end{minipage}
\quad
{or}
\quad
\begin{minipage}{0.25\textwidth}
\centering
\emph{(d). -100, 10. 0, 0.\\
(d). -100, 4. 0, 0.\\
(). -60, 0. 0, 0.\\
(). -4, 0. 0, 1.}
\end{minipage}
\end{center}
where the string includes: \texttt{(keyboard inputs). horizontal mouse move, vertical mouse move. left mouse click, right mouse click.} 
Example keyboard inputs are (w, a, s, d, space, ...) -- walk forward, left, backward, right, jump, etc. See \citet{pearce2022csgo} for additional details on the \CSGO action space. 

In the case of \name, the meta-prompt text is inserted at the beginning of the sequence, and the controller text is inserted just before the associated block of frames. In the case of \ThinktoV, all controller actions are inserted at the beginning of the sequence prior to any frames, along with the meta-prompt. In the case of \TtoV, no controller actions are provided. Again, refer to \autoref{fig:csgo_data} for a visualization. 

The \CSGO dataset has a resolution of $280 \times 150$ at 16 FPS, which we upsample to $320 \times 192$ for tokenizer compatibility. 
With a $2\times2$ U-Net patching and $4\times8\times8$ tokenizer, each chunk of 4 frames amounts to 240 tokens. 
The models' context size fits 6.1 seconds (or 98 frames) of video per step per device during model training.

\subsection{Evaluation set-up}\label{subsec:csgo_eval_setup}

We evaluate the overall video quality and user controllability of videos generated using the \name approach with in-house blind human annotations.

\paragraph{Evaluating overall video quality}
To test the hypothesis that interleaved planning text improves video generation quality, we perform pairwise comparisons between videos generated by \name and those from the non-interleaved baselines \TtoV and \ThinktoV.

For all models, we generate 100 short-form and 40 long-form videos, yielding 280 pairwise annotation tasks. 
Short-form videos are approximately six seconds long, and long-form videos are 64 seconds long (see \autoref{sec:method_inference} for details on how we generate long videos).
For a given model, each generated video is conditioned on a single, unique frame from a held-out expert gameplay set and the prompt: ``\emph{A brief gameplay clip from the iconic Dust 2 map, showcasing classic Counter-Strike tactical combat in the legendary desert setting.}''
Videos evaluated side-by-side for visual quality share the same conditioning frames. 
Distortions and cloudiness in generated videos are penalized, along with physically implausible behaviors such as wall-clipping, abnormally slow gameplay, or spontaneous teleportation.  
Exact evaluation criteria can be found in Appendix~\ref{app:csgo_eval}.

\paragraph{Evaluating fine-grained controllability}
To test the hypothesis that interleaved planning text affords strong, fine-grained controllability relative to non-interleaved planning, we compare \name's response to intermediate text interventions against the \ThinktoV baseline.

For \name, we generate videos similar to those described for the overall video quality evaluation but with an additional invocation of a manual intervention (\eg a ``left-click'' action, a ``reloading'' action, etc.) at an intermediate timestamp of the video. 
We experiment with interventions at $t=0.8125s$ and $2.8125s$, which we refer to as 1s and 3s for simplicity. 
We compare against videos generated by the \ThinktoV model, where the intervention is inserted into the model-generated text plan used as initial conditioning for video generation.
To isolate the effect of these interventions, we include control videos from both models generated from the same conditioning frames and meta-prompt \textit{without} any manual intervention.

In total, four user-controllable text interventions are evaluated: moving backwards, performing a ``left-click'' action that corresponds to shooting a weapon, initiating a weapon reload, and jumping. 
Control videos with no manual intervention are also included, denoted as ``no-op'' (no operation) interventions, to provide insight into how often an action is generated by the model even without manual intervention.
For each model, we evaluate 150 generated videos with interventions, yielding 300 annotation tasks. 

Human annotators provide point-wise evaluations of: 
\begin{enumerate}
    \item \textbf{Intervention Correctness}, i.e. how well the generated video reflects the user-specified intervention. Users select among three options: the video correctly reflects the intervention ($+1$), incorrectly reflects the intervention ($0$), or that they are unsure ($+0.5$). (Conversely, we mark \emph{no-op} control videos as correct when the annotator marks that the action is not reflected in the video.) We average these results to obtain a score.  
    \item \textbf{Visual Quality}, using the same quality criteria defined above. Users have the option to rate the visual quality as strong ($+3$), moderate ($+2$), weak ($+1$), or none ($0$). We average these results and normalize to obtain a score between $0$ and $1$.
\end{enumerate}
The exact evaluation instructions are provided in the Appendix \autoref{app:csgo_eval}.

\subsection{Results}

\begin{figure}
    \centering
    \includegraphics[width=\textwidth]{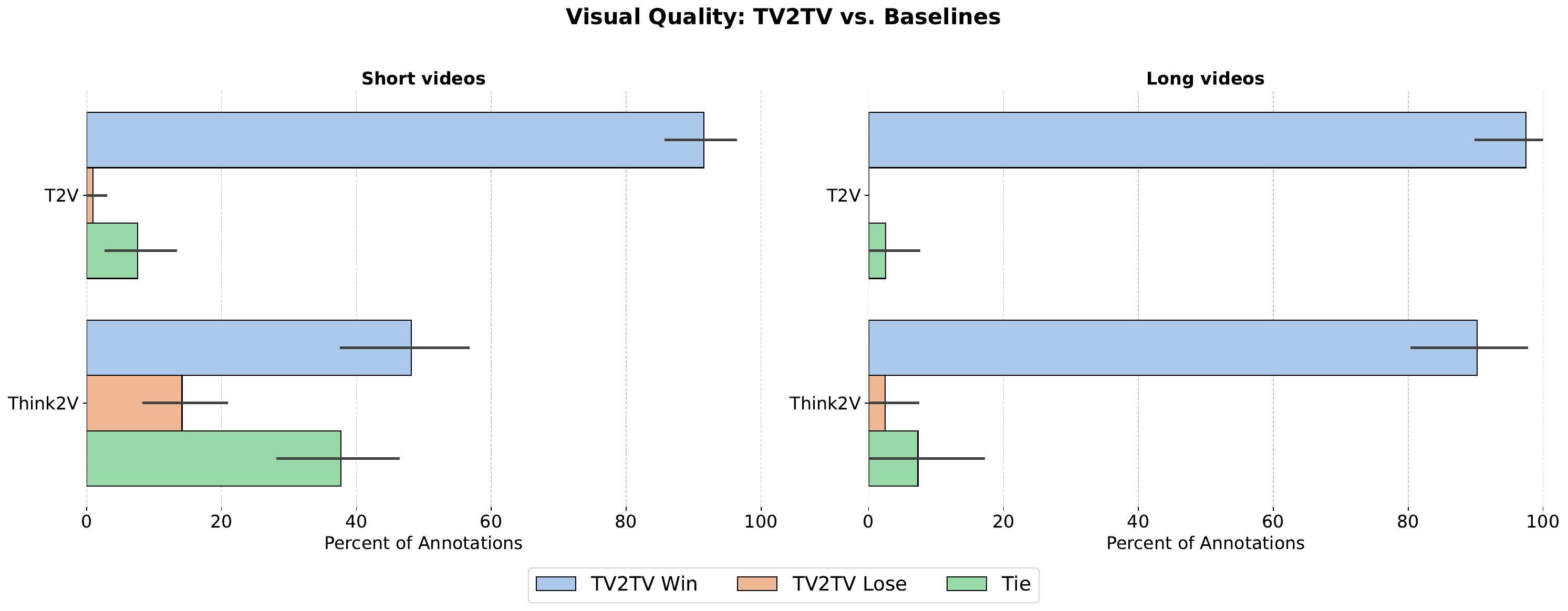}
    \caption{\textbf{Visual quality human evaluation results.} Pairwise visual quality comparison of videos generated by \name, \TtoV, and \ThinktoV. Results are shown for our standard 6-second context length (short videos) and extended 64-second videos (long videos) generated using sliding windows (see \autoref{sec:method_inference}). 95\% confidence intervals are shown. In both settings, \name outperforms baselines.
    }\label{fig:rq1_results}
\end{figure}

\begin{figure}[h]
  \centering
  \includegraphics[width=0.9\linewidth]{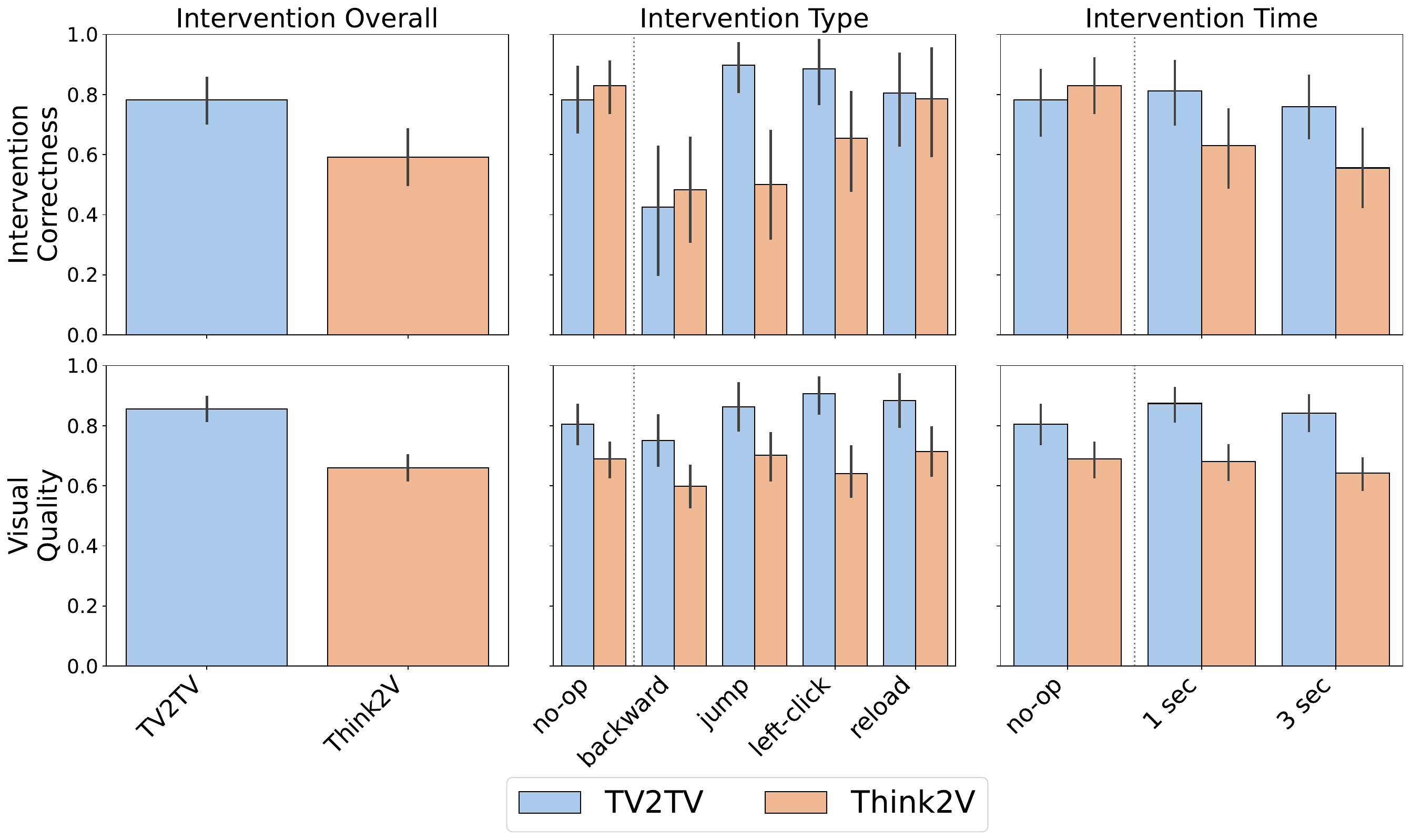}

  \caption{\textbf{Fine-grained controllability human evaluation results.} Pointwise controllability (Intervention Correctness) and quality (Visual Quality) comparison of videos generated by \name and \ThinktoV with intermediate action interventions. Results show that \name demonstrates significant controllability advantages over \ThinktoV with particularly strong control for \emph{jump} and \emph{left-click} and similar control for \emph{reload} and \emph{backward}.  Moreover, \name produces signficantly higher visual quality generations with these interventions. 
  Both scores range from 0 to 1. 
  We also evaluate the \emph{no-op} (no operation) baseline for reference -- in the case of \emph{no-op}, the video is marked as correct if the action is \emph{not shown} in the video. We find that, when not manually intervened on, models generate the action only 17-22\% of the time, showing that the intervention correctness during manual intervention is significantly above random.
  Details on metrics are provided in \autoref{subsec:csgo_eval_setup}.
  }
  \label{fig:csgo_rq2}
\end{figure}

Results for overall visual quality and controllability experiments are shown in \autoref{fig:rq1_results} and  \autoref{fig:csgo_rq2}, respectively. We make the following observations:

\begin{itemize}

    \item{\textbf{Generating interleaved planning text substantially improves overall video quality.}} 
    As shown in \autoref{fig:rq1_results}, \name achieves substantial improvements in overall visual quality compared to both \TtoV and \ThinktoV across short and long videos. The performance gains are more substantial relative to \TtoV (91\% \textit{vs.} 1\%, with 8\% ties) than to \ThinktoV (48\% \textit{vs.} 14\%, with 38\% ties). Additionally, improvements are more pronounced for long videos compared to short videos across both baseline comparisons, which underscores \name's effectiveness for long-form video generation.

    \item{\textbf{Interleaved text affords strong, fine-grained controllability over video generation.}} 
    \autoref{fig:csgo_rq2} shows intervention correctness and visual quality results for four types of intermediate interventions: firing (\emph{left-click}), jumping (\emph{jump}), reloading (\emph{reload}), and moving backward (\emph{backward}), each applied at timestamps $t=0.8125s$ and $2.8125s$, denoted 1s and 3s for simplicity. 
    \name demonstrates significant controllability advantages over \ThinktoV (78\% \textit{vs.} 59\%), with particularly strong control for  \emph{jump} and \emph{left-click} interventions. Controllability is similar betwen the two models for \emph{reload} and \emph{backward}. 
    \name shows stronger intervention controllability than \ThinktoV for both intervention timestamps.
    For reference, we also evaluate the \emph{no-op} (no operation) baseline, where no manual intervention is applied -- in the case of \emph{no-op}, the video is marked as correct if the action is \emph{not shown} in the video. We find that, when not manually intervened on, models generate the action only 17-22\% of the time. This highlights that the intervention correctness during manual intervention is significantly above random.
    Additionally, \name produces significantly higher visual quality generations during manual interventions than \ThinktoV. 

\end{itemize}

%% file: sections/05_realworld.tex
\section{Scaling \name to Real World Data}\label{sec:soccer}

While video games offer a convenient source of interleaved text-and-video data, most real-world video data lacks clean, temporally-aligned (timed), interleaving captions, presenting a challenge for extending \name to open domain settings.  In this section, following recent work on dense, differential video captioning \citep{chen2024sharegpt4video, chen2024makes, peng2025patch, chen2025planning, wang2023internvid, xue2025progress}, we present our methodology for developing a data augmentation pipeline that synthetically augments real-world videos with interleaved natural language action descriptions using vision-language models (VLMs). This enables us to extend and generalize \name to large-scale real-world data. We apply this pipeline to \emph{sports videos} -- chosen for their dynamic motion and rich action content -- to create a dataset of 8K hours of interleaved text-and-video data. Finally, we evaluate \name trained on this augmented data from scratch against established video generation models\footnote{Sports highlights contain fast motion and relatively complex semantics. This makes them a challenging test case for existing pretrained video generation models, where sports content is present but not well represented in the training distribution.} as well as controlled baselines.

\subsection{Interleaved data augmentation pipeline}
\label{sec:data-pipeline}

\begin{figure}
  \centering
    \includegraphics[width=0.82\columnwidth]{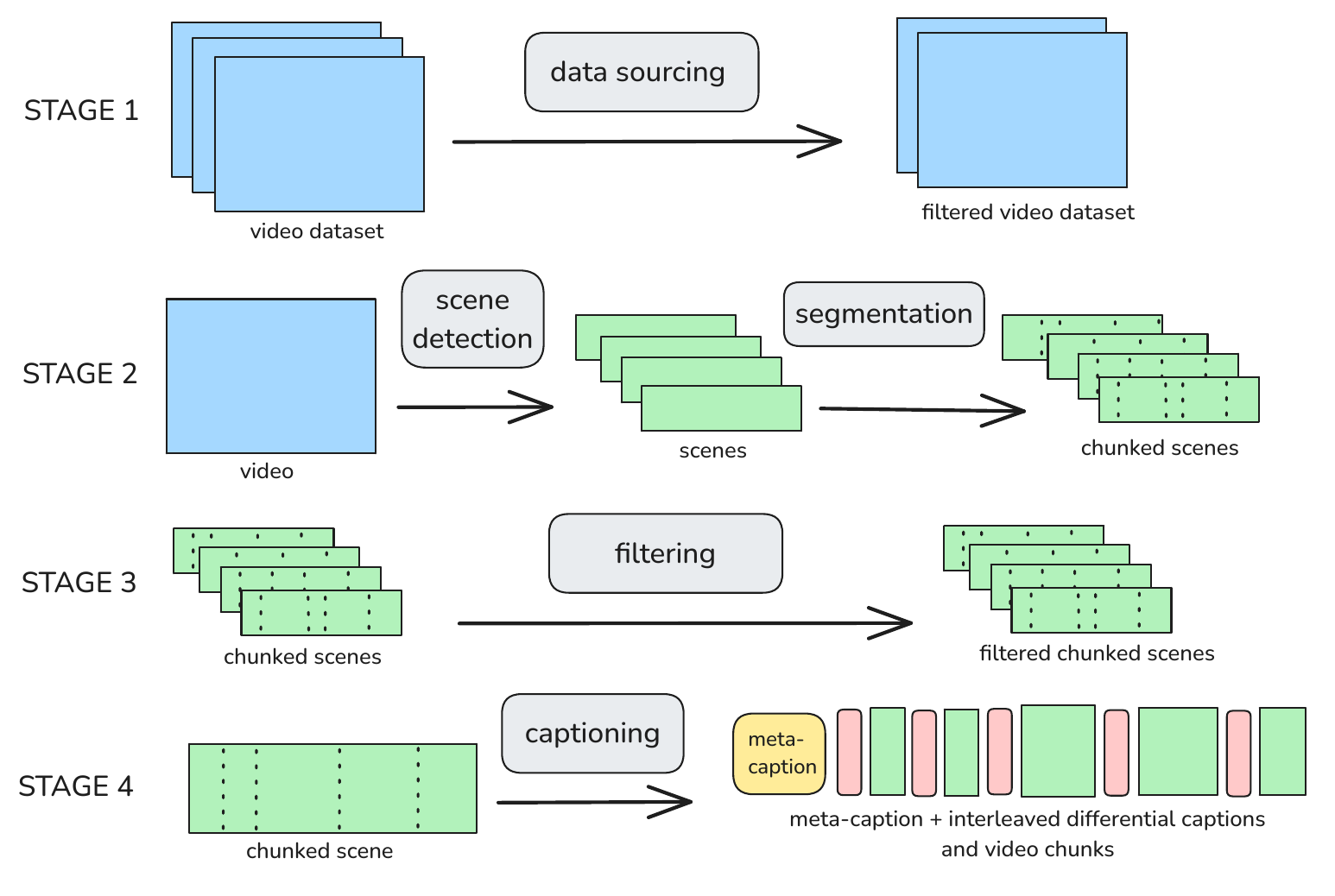}
    \caption{\textbf{Interleaved text and video data pipeline}. We present our methodology for using constructing an interleaved text and video training dataset from real world videos. The pipeline consists of several stages, including (1) data sourcing, (2) scene detection and segmentation, (3) filtering, and finally, (4) interleaved captioning with a VLM.}
\label{fig:data_pipeline}
\end{figure}

We describe how we construct interleaved text-and-video data from videos. The pipeline consists of several stages:
\begin{enumerate}
\item \textbf{Data sourcing:} Collecting sports domain data from the large-scale YT-Temporal-1B \citep{zellers2022merlotreserve} dataset, using keyword-based and other filtering criteria.
\item \textbf{Scene detection and segmentation:} Dividing videos into 6 to 16-second scenes, and further sub-dividing those scenes into segments based on detected key frames and other heuristics. 
\item \textbf{Filtering:} Selecting high-quality scenes using model-based and other filters that assess motion characteristics, semantic content, and overall quality. 
\item \textbf{Interleaved captioning:} Generating captions at multiple levels using a VLM: an overall meta-caption for each scene and fine-grained, interleaved captions describing changes between short segments. 
\end{enumerate}
See \autoref{fig:data_pipeline} for a visualization of this pipeline.

\paragraph{Data sourcing} We focus on building a dataset of sports content by filtering the YT-Temporal-1B dataset \citep{zellers2022merlotreserve} using keyword-based filters (e.g. ``\emph{game highlights}''). We chose the sports domain for its high action density, which provides a strong testbed for interleaved reasoning capabilities. This yields 38K total hours of data. 

\paragraph{Scene detection and segmentation} We segment the data into scenes using TransNetV2 \citep{soucek2020transnetv2}, a shot boundary detection model based on 3D convolutional networks. To identify clips containing interesting content, we employ a two-step approach. First, we use the Perception Encoder \citep{bolya2025perceptionencoder} to embed video frames and compute the cosine distance between consecutive frames, producing a time-series of semantic change. Peaks in this time-series may indicate moments where significant action likely occurs, so we refer to the frames associated with these peaks as \emph{key frames}. We then apply an 8-second sliding window to each scene, extracting clips that are between 6 and 16 seconds long and contain the highest number of peaks. On average, these clips are 8.2 seconds in length.

Finally, we use a combination of key frames, hierarchical clustering of Perception Encoder embeddings \citep{chen2025planning}, and heuristics to further segment the clips into chunks of frames suitable for interleaved captioning. Each clip is divided into an average of 4.3 chunks, though the number of chunks can range from as few as 2 to as many as 10 depending on the length of the video and the number of key frames detected. We impose a minimum chunk length of 1-second; the average chunk length is 1.9
seconds.
 
\paragraph{Filtering} We apply several scene-level filters to further refine our selection:
\begin{itemize}

\item \textbf{VLM-based quality classifier:} To select high quality scenes we prompt Gemma-3-12B-Instruct \citep{gemma3} to select semantically-relevant content. For each scene, we sample consecutive frames from the start, center, and end of the video and ask the model to provide a score of 1-10 based on the perceived quality and relevance; see \autoref{appendix:prompt} for the full prompt. 

\item \textbf{Face bounding box filter:} We observed that a considerable portion of videos consisted of people talking directly to the camera without meaningful action or motion in the foreground. To remove such videos, we use RetinaFace \citep{deng2019retinaface} to obtain face bounding boxes and analyze both their coverage and temporal stability throughout the video. Clips with large, stable face bounding boxes are filtered out.

\item \textbf{Motion filter:} We compute the optical flow for each clip \citep{farneback2003opticalflow} and calculate its average magnitude across frames as a motion score. Clips with low motion scores, indicating static or minimal movement, are filtered out. 

\end{itemize}
After filtering, our final dataset comprises 8K hours of sports video data.

\paragraph{Interleaved captioning} 
Finally, following recent work on differential video captioning \citep{chen2024sharegpt4video, chen2024makes, peng2025patch, chen2025planning, wang2023internvid, xue2025progress}, we use Qwen3-VL-30B-A3B-Instruct \citep{Qwen2.5-VL} 
to generate (1) an overall \emph{meta-caption} for the video and (2) differential captions describing action changes across subsequent frame chunks.
Detailed prompts passed to the VLMs are provided in the Appendix \autoref{appendix:differential_caption_prompt}.
An example interleaved document produced by this pipeline is provided in \autoref{tab:video_frames}.

\subsection{Experiments and analysis}

\subsubsection{Modeling and data details}

We adopt an 8B-MoT backbone with modality-specific 8B-parameter text and video towers (\autoref{sec:methods}). The text tower is initialized with Llama-3.1-8B \citep{llama3}. We train for 250K steps utilizing a batch size of 512 and a cosine learning rate scheduler with a maximum learning rate of $3 \mathrm{e}{-4}$.  Detailed model and training configurations are provided in \autoref{tab:model_configs_sports} in \autoref{appendix:configuration_sports}. 

We downsample the video data to $320 \times 192$ resolution at 16 FPS and randomly sample 6.1 seconds (or 98 frames) of video per step per device for model training.

{% Begin local group
\setlength{\aboverulesep}{0pt}
\setlength{\belowrulesep}{0pt}
\newcolumntype{P}[1]{>{\raggedright\arraybackslash\vspace{-1ex}}p{#1}}
\begin{table}[t]
\centering
\small
\setlength{\tabcolsep}{3pt}
\begin{tabular}{@{}P{0.4\textwidth}P{0.58\textwidth}@{}}
\toprule
\textbf{Interleaved caption} & \textbf{Subsampled frames} \\
\midrule
\textbf{[0s - 2.2s]} {\footnotesize A player in a white uniform with pink socks runs with the ball, evading a defender in a black uniform. The player in white sprints towards the sideline, maintaining possession of the ball. The defender in black trails behind, attempting to catch up. Spectators are visible in the background.} & 
\includegraphics[width=0.18\textwidth]{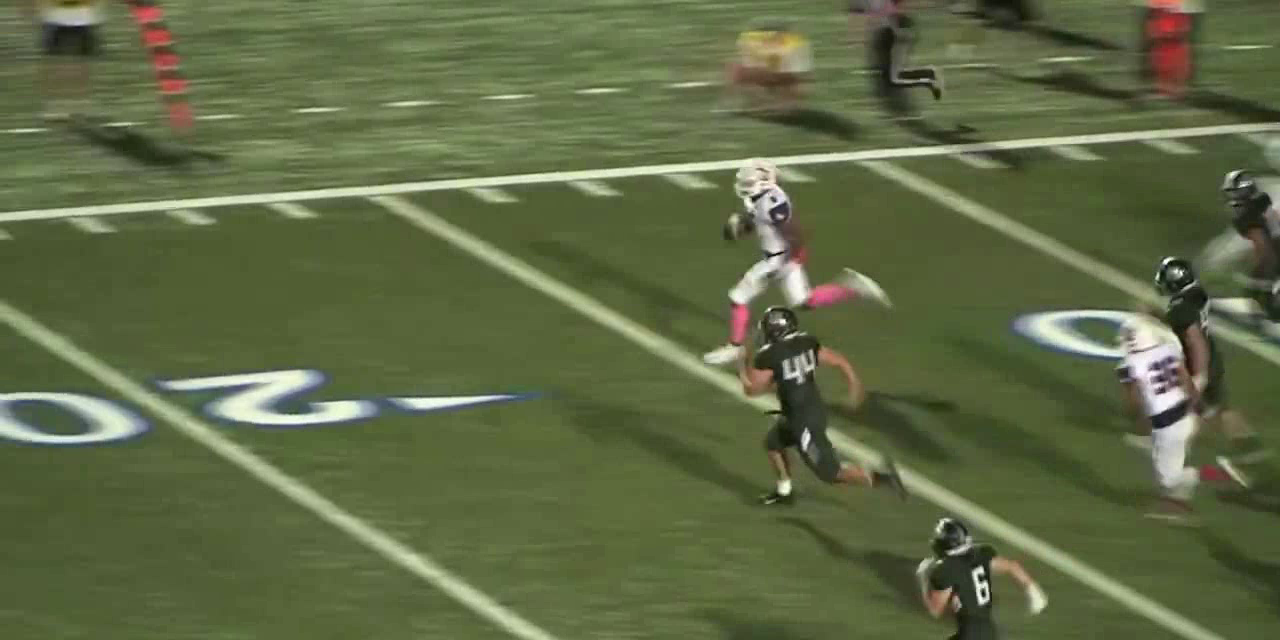}
\includegraphics[width=0.18\textwidth]{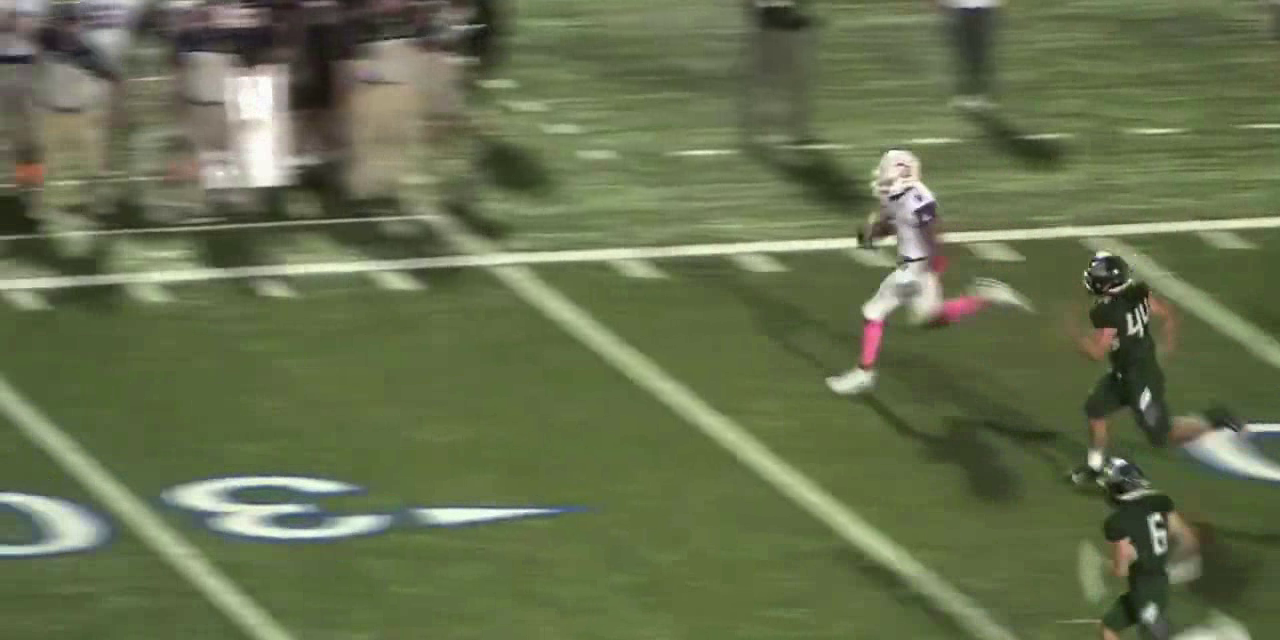}
\includegraphics[width=0.18\textwidth]{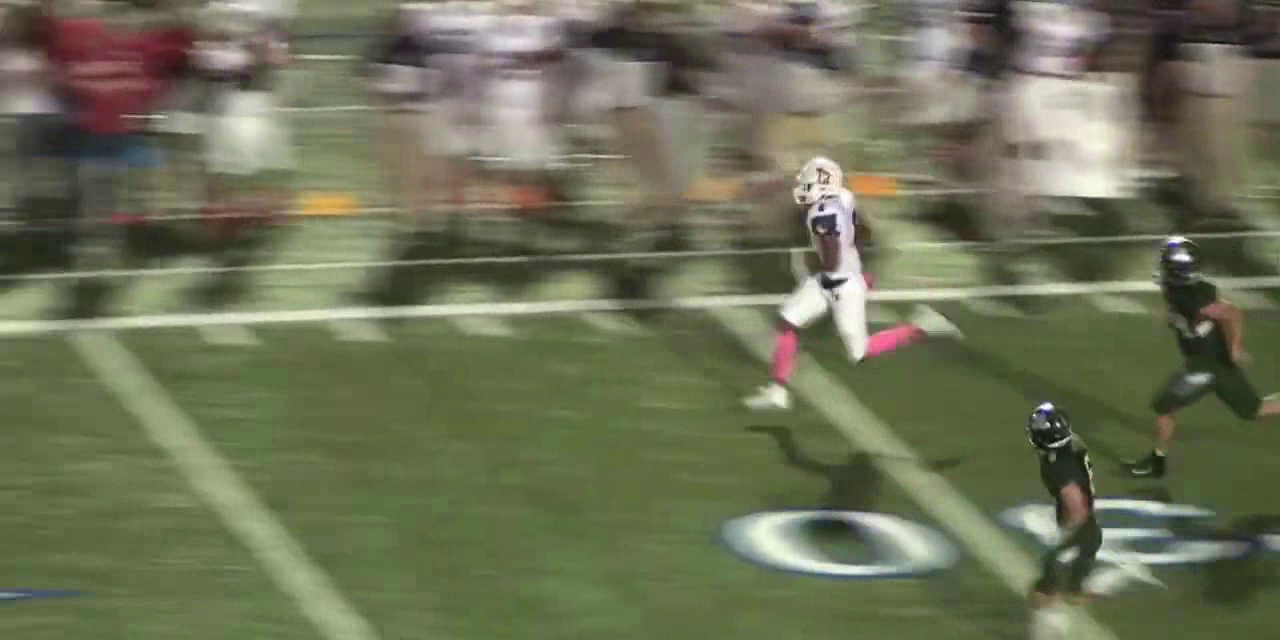} \\
\midrule
\textbf{[2.2s - 4.3s]} {\footnotesize The player in the white uniform with pink socks continues running with the ball, moving further downfield. The defender in the black uniform remains in pursuit, closing the distance. Additional players in black uniforms join the chase, running towards the player with the ball. The player in white maintains possession and speed, evading the approaching defenders.} & 
\includegraphics[width=0.18\textwidth]{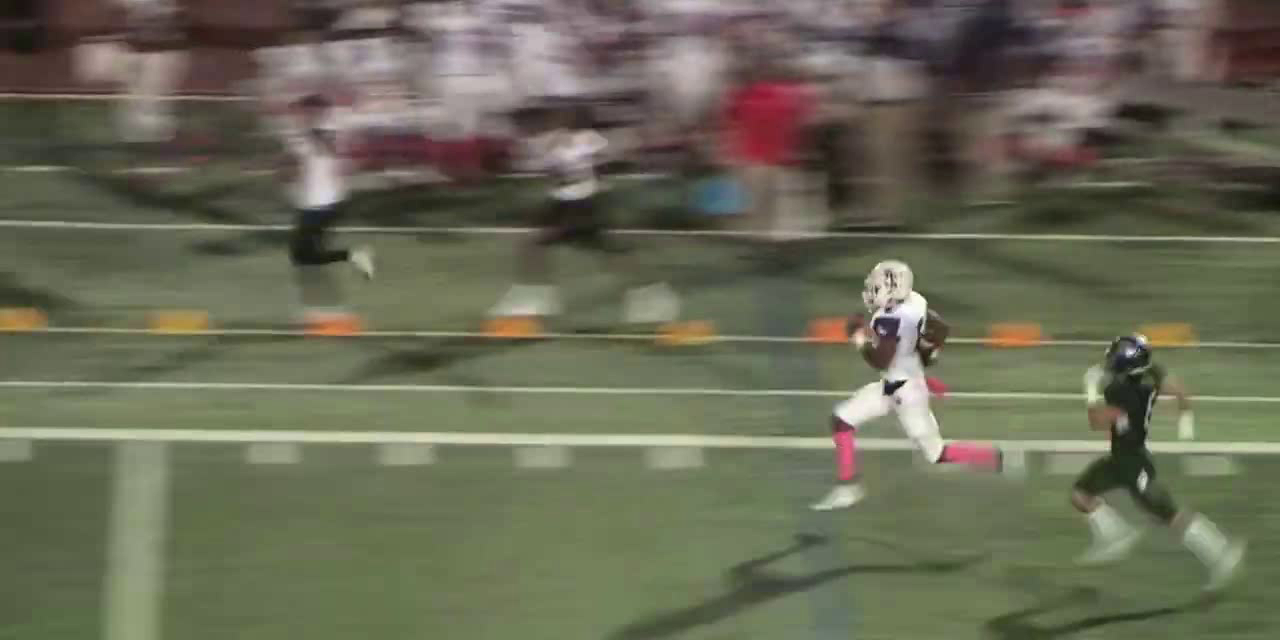}
\includegraphics[width=0.18\textwidth]{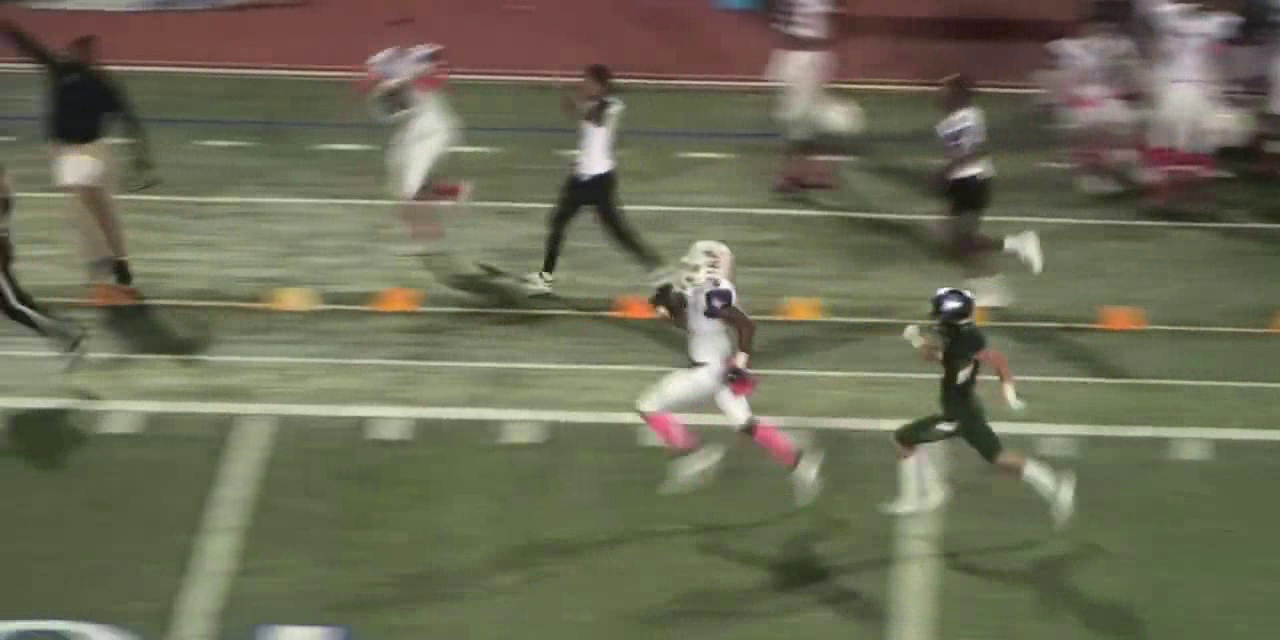}
\includegraphics[width=0.18\textwidth]{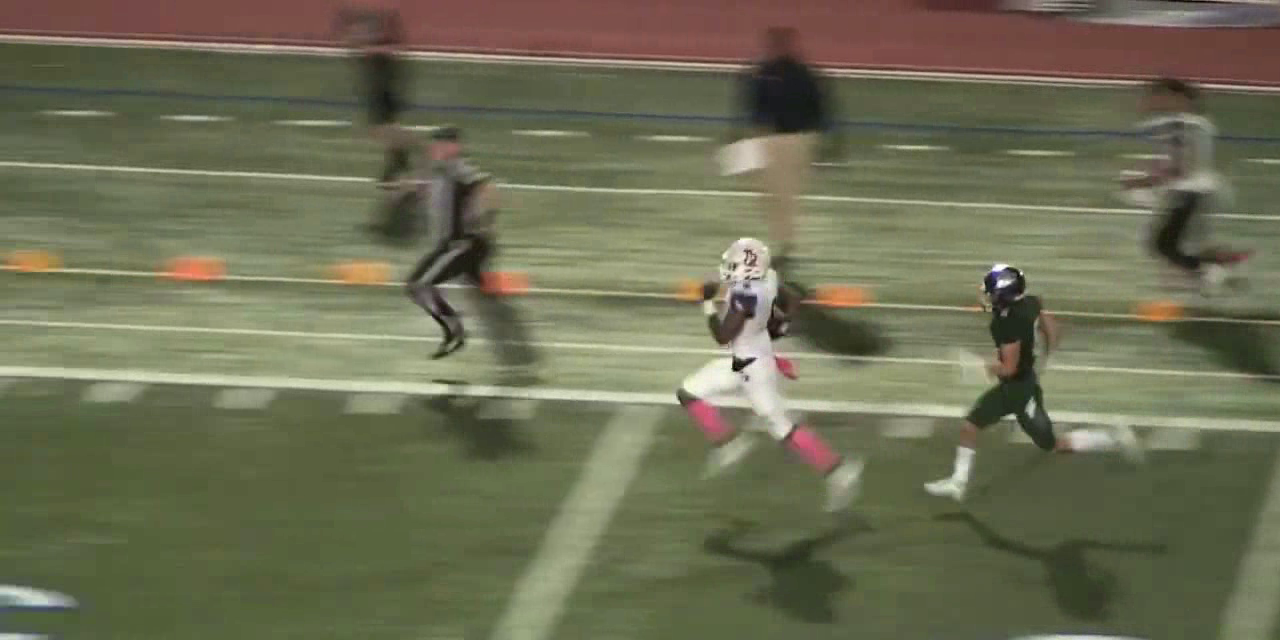} \\
\midrule 
\textbf{[4.3s - 6.0s]} {\footnotesize The defender in the black uniform attempts a tackle but misses, falling to the ground. The player in the white uniform continues running unopposed towards the end zone.}  & 
\includegraphics[width=0.18\textwidth]{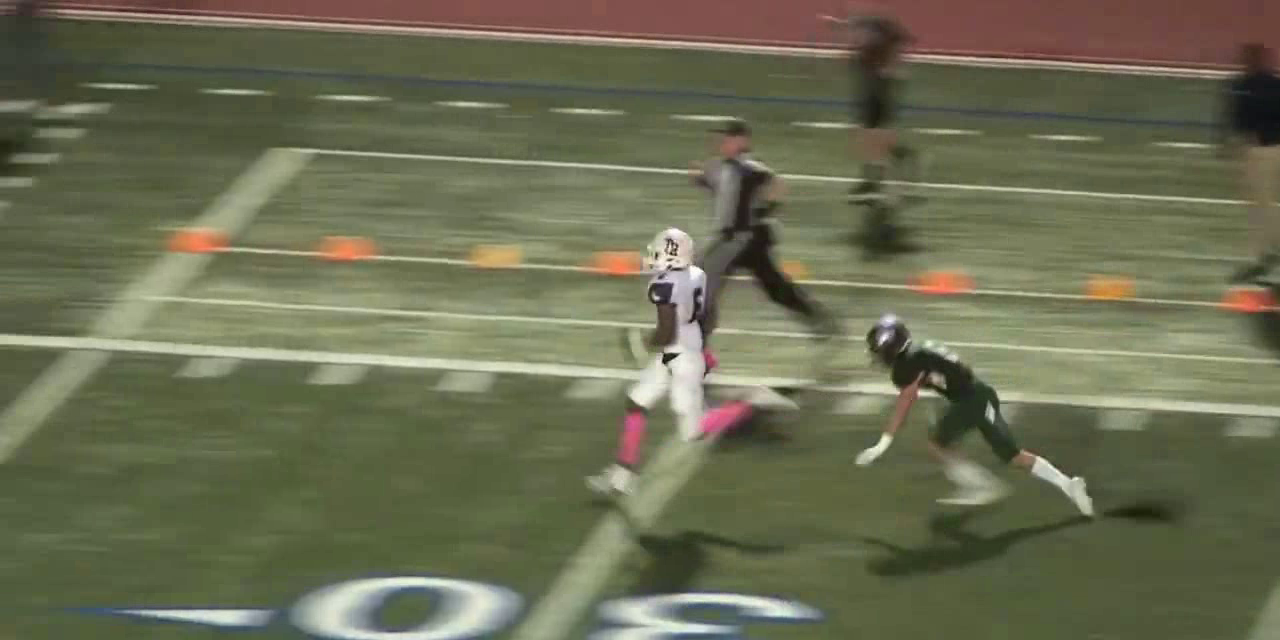}
\includegraphics[width=0.18\textwidth]{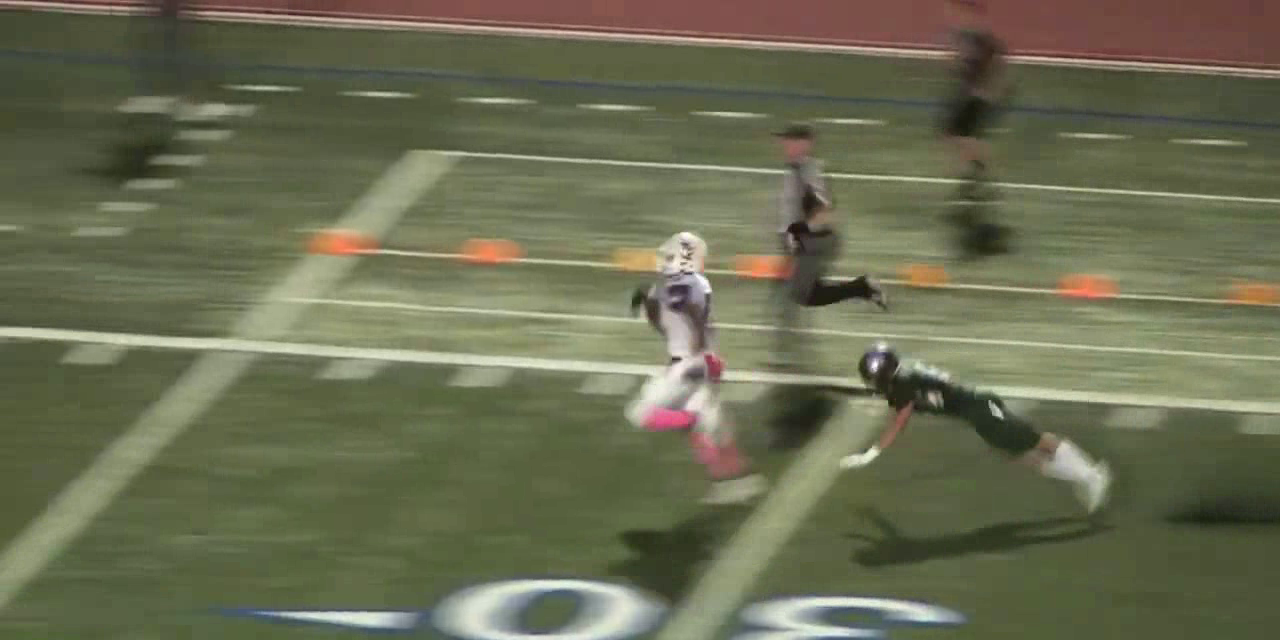}
\includegraphics[width=0.18\textwidth]{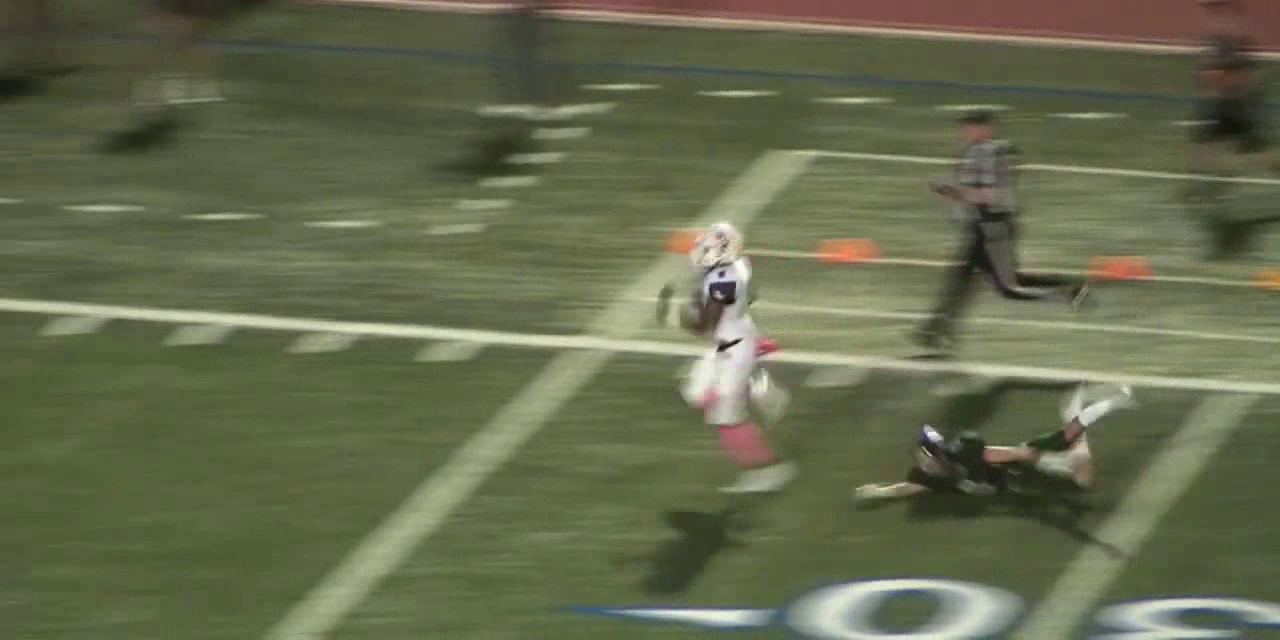} \\
\midrule 
\textbf{[6.0s - 8.6s]} {\footnotesize The player in the white uniform continues running towards the end zone, approaching the goal line.} & 
\includegraphics[width=0.18\textwidth]{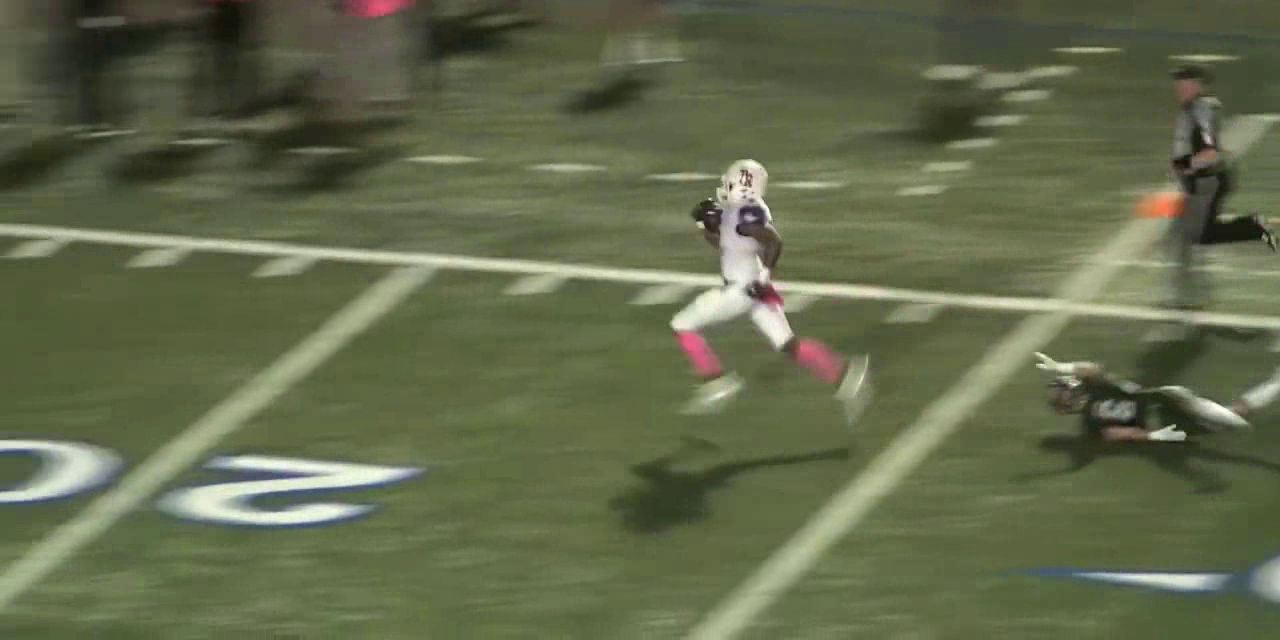}
\includegraphics[width=0.18\textwidth]{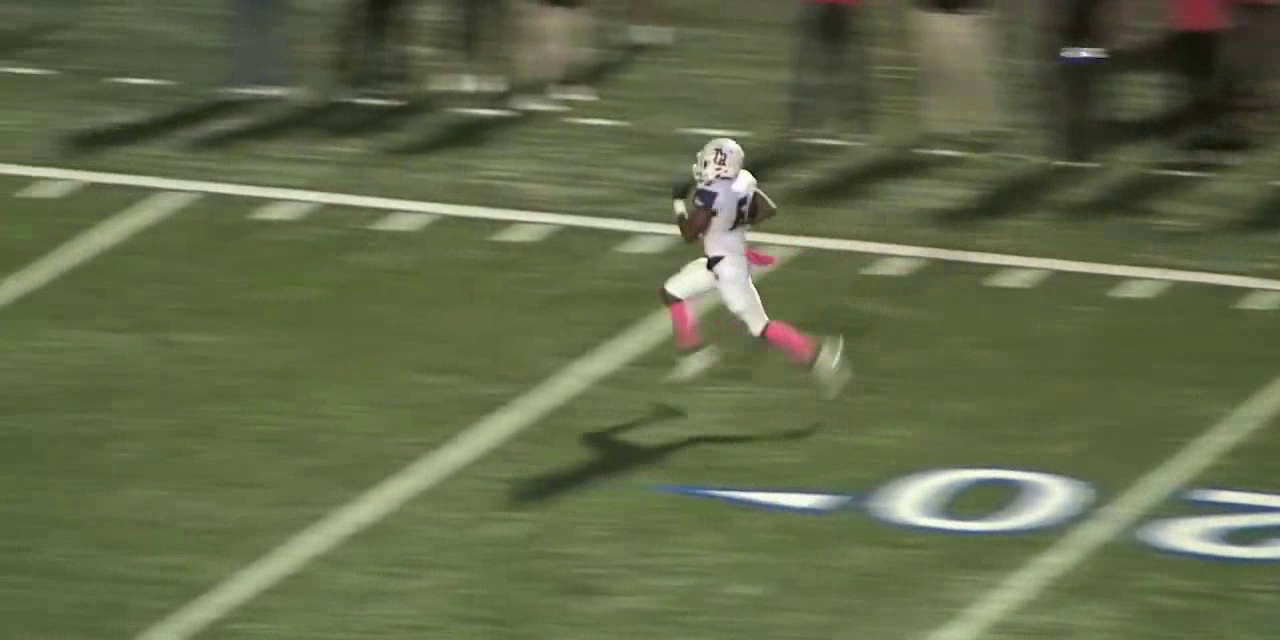}
\includegraphics[width=0.18\textwidth]{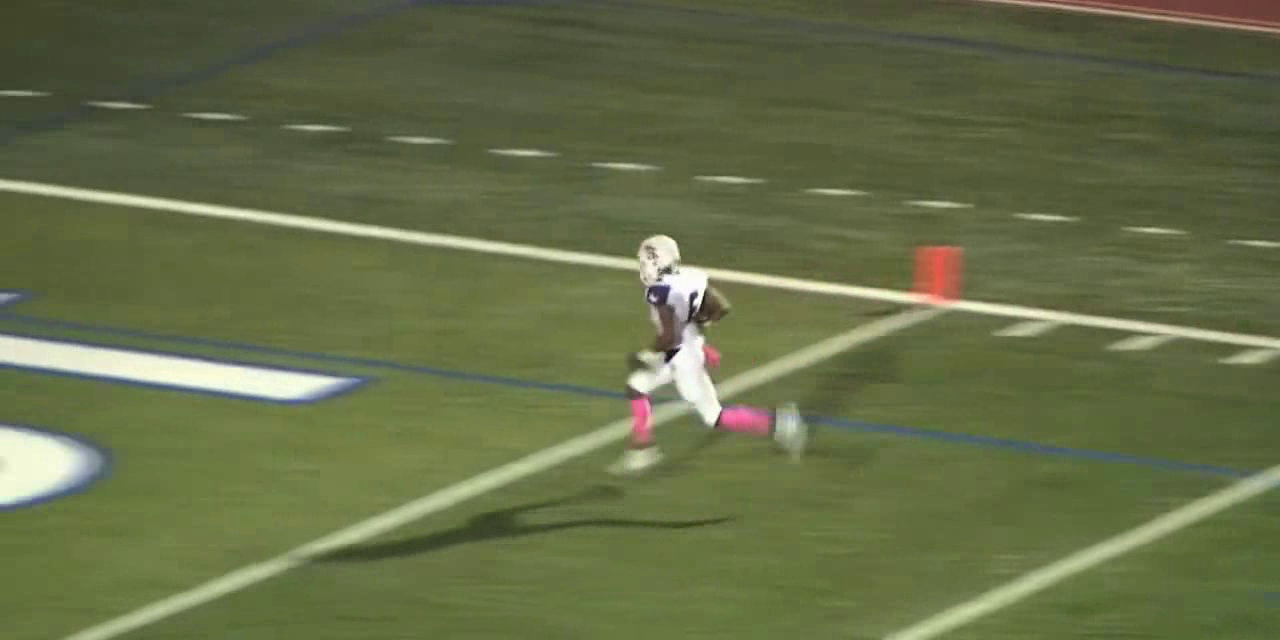} \\
\midrule 
\textbf{[8.6s - 10.1s]} {\footnotesize The player in the white uniform crosses the goal line, scoring a touchdown. The ball is now on the ground near the end zone.}
 & 
\includegraphics[width=0.18\textwidth]{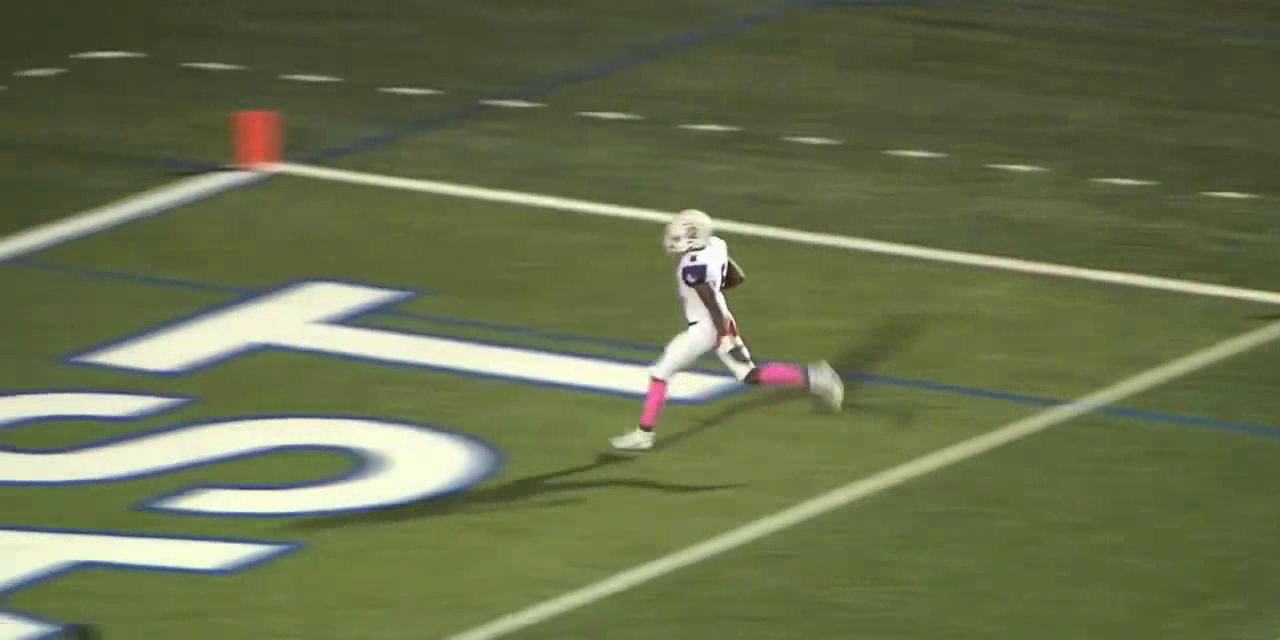}
\includegraphics[width=0.18\textwidth]{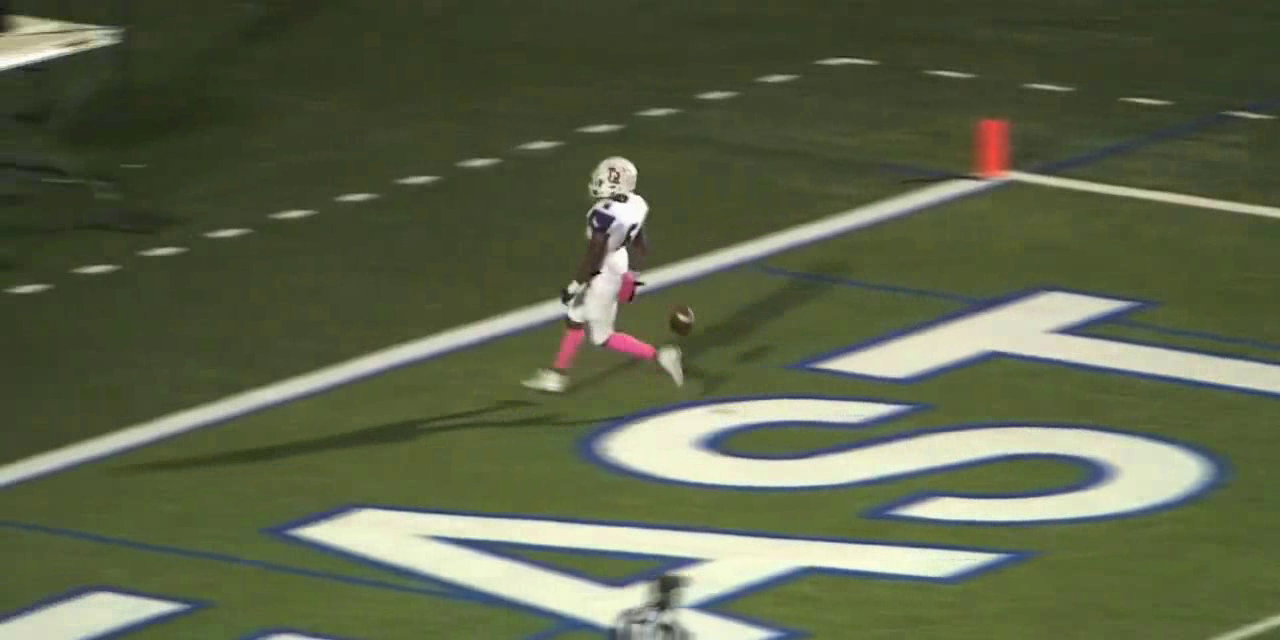}
\includegraphics[width=0.18\textwidth]{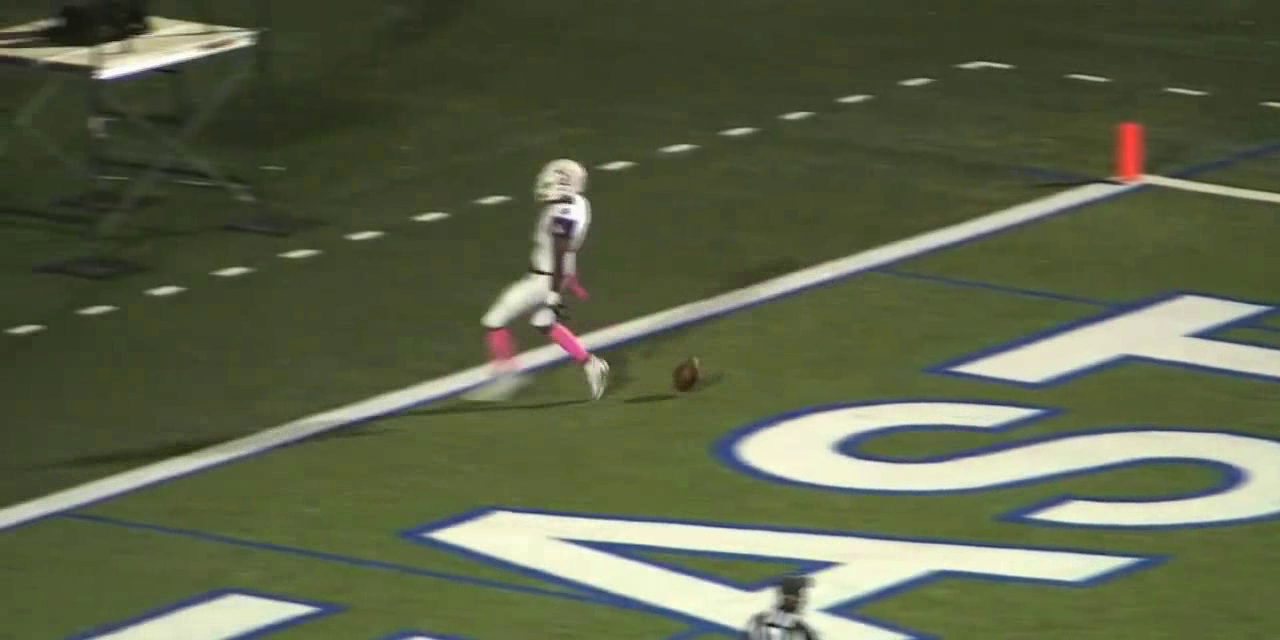} \\
\midrule 
\addlinespace[2pt]  % Add 6pt of space
\multicolumn{2}{@{}p{0.95\textwidth}@{}}{\textbf{Overall meta-caption:} A player in a white uniform with pink socks runs with the ball, evading defenders, and scores a touchdown.} \\
\addlinespace[2pt]  % Add 6pt of space
\bottomrule
\end{tabular}
\caption{\textbf{Example interleaved training document}. The source data is from YT-Temporal-1B \citep{zellers2022merlotreserve}.\protect\footnotemark \xspace 
Interleaved captions and the overall meta-caption are generated by Qwen3-VL-30B-A3B-Instruct~\citep{Qwen2.5-VL}.}
\label{tab:video_frames}
\end{table}
}% End local group - settings don't affect anything else
\footnotetext{Original video source: \url{https://www.youtube.com/watch?v=v6htOcLa7KM}}

\subsubsection{Evaluation set-up} 

We conduct pairwise evaluations comparing videos generated with \name and (1) established, external video generation models and (2) controlled \TtoV and \ThinktoV baselines. 
Pairs are evaluated by a pool of professional external annotators via the Turing platform for increased robustness.

We curate a held-out sports evaluation set consisting of major sports highlights and use meta-prompts captioned by VLMs as text prompts. 
The first frame of each video is used as the initial image condition. 
For each pairwise comparison, annotators assess prompt alignment, visual quality, real-world fidelity, and overall preference.
Evaluation instructions used by annotators can be found in the Appendix~\autoref{appendix:configuration_sports}.

\begin{figure}[t]
  \centering
  \begin{subfigure}{0.77\columnwidth}
    \centering
    \includegraphics[width=\linewidth]{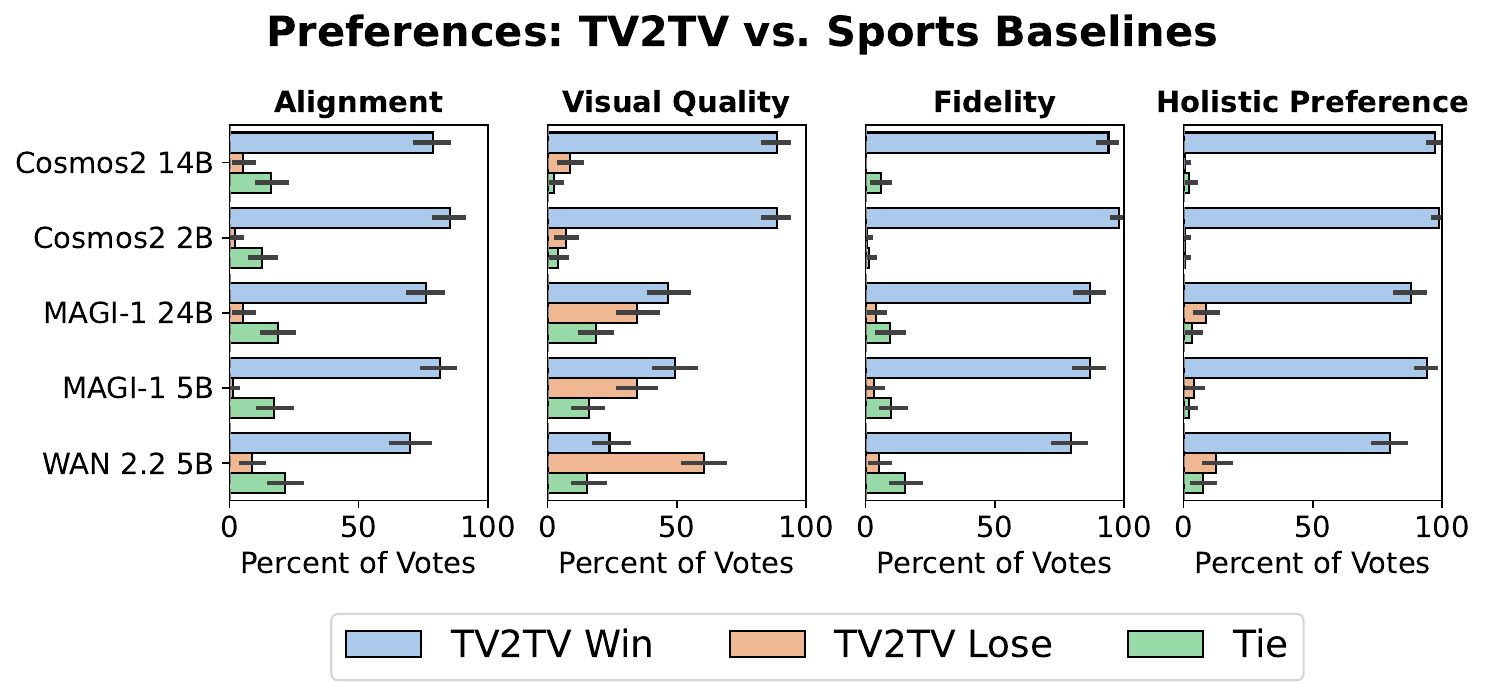}
  \end{subfigure}

  \caption{\textbf{Evaluation of \name trained on interleaved sports data vs. external video models.} We note that the external models we compare to were not specifically tuned for the sports domain and as such we do not expect them to perform well; however, we include them as out-of-domain baselines to obtain insights into \name's visual quality and prompt alignment relative to high-performing general-purpose models in this challenging domain. In blind pairwise human evaluations by external annotators, \name outperforms all models on sports prompts in prompt alignment, real-world fidelity, and overall holistic preference.  For visual quality, we find that \name surpasses Cosmos2 variants, has similar performance as MAGI-1 variants, and underperforms compared to WAN 2.2 5B. 95\% confidence intervals are shown.}
  \label{fig:sports_rq1}
\end{figure}

\paragraph{External models} We compare videos generated with \name to those produced by several established video generation models: Cosmos-Predict2-Video2World 2B and 14B variants \citep{agarwal2025cosmos}, MAGI-1 4.5 and 24B variants \citep{teng2025magi}, and WAN-2.2 TI2V 5B \citep{wan2025}. 
All of these models include a TI2V mode, which conditions video generation on a text prompt and an initial frame. 
We note that the external models we compare to were not specifically tuned for the sports domain and as such we do not expect them to perform well; however, we include them as out-of-domain baselines to obtain insights into \name's visual quality and prompt alignment relative to high-performing general-purpose models in this challenging domain. 

For each model, we generate five videos for each of 30 unique conditioning images and prompts, yielding a total of 750 pairwise comparison tasks between \name and external models. 
For the external models, we generate videos in their native resolutions. 
To allow for a balanced comparison, prior to human evaluation stage we downsample the videos generated with external models to the shared lowest resolution across all models ($320 \times 192$) and FPS ($16$) of \name. 

\paragraph{Controlled baselines}  We compare videos generated with \name to those generated by \TtoV and \ThinktoV baselines trained under the same settings (but varying the data representation for each framework). 
Similar to \autoref{subsec:thinktov_ttov}, \TtoV is trained without any interleaved text, conditioning directly on a concise meta-prompt. \ThinktoV is trained on a concatenation of the meta-prompt and an extended, detailed prompt. Compared to the meta-prompt, the extended prompt explicitly instructs VLMs (Qwen3-VL-30B-A3B-Instruct) to provide more thorough and detailed descriptions of objects, actions, event progression, and so on, based on more densely sampled video frames. During inference, \ThinktoV first self-generates (thinks) such detailed descriptions conditioned on the meta-prompt, and then generates video frames in a non-interleaved manner. This setup is similar to the use of highly descriptive synthetic captions in \citet{betker2023improving}.  

For each model, we generate five videos for each of 30 unique conditioning images and prompts, yielding a total of 300 pairwise comparison tasks between \name and these baselines.

\begin{figure}[t]
    \centering
    \includegraphics[width=0.77\columnwidth]{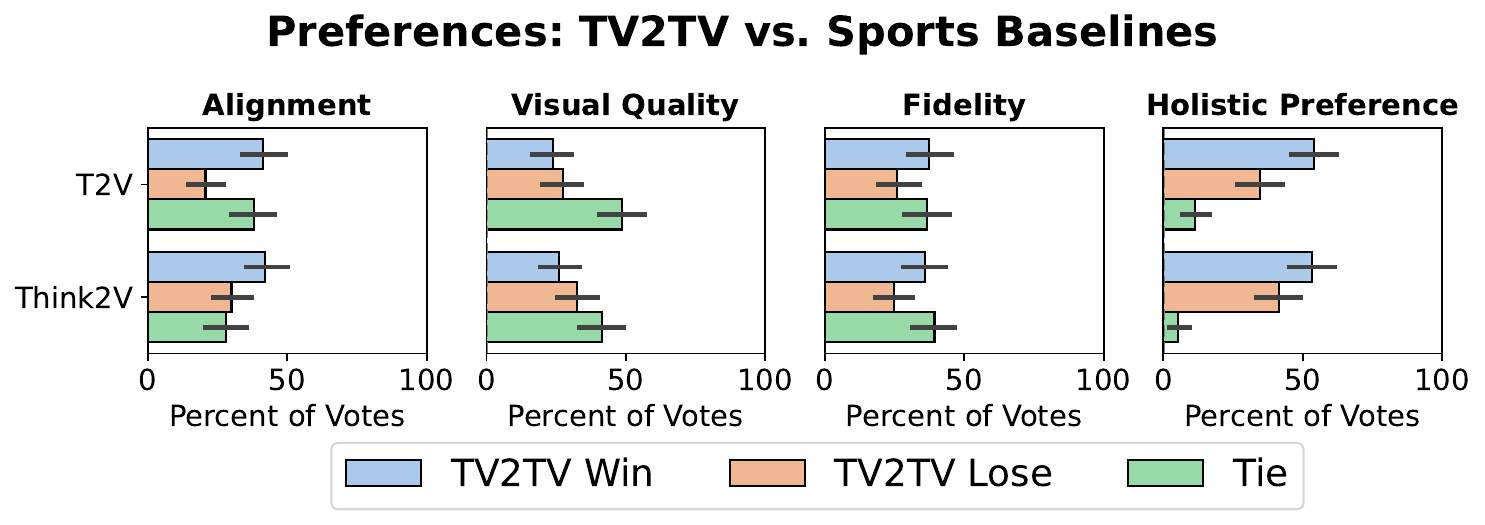}
    \caption{\textbf{Evaluation of \name trained on interleaved sports data \emph{vs.} \TtoV and \ThinktoV in a controlled setup.} Compared to \TtoV, \name shows stronger alignment and holistic preference, with similar visual quality and fidelity. When compared to \ThinktoV, \name shows a similar pattern of improved alignment and overall preference, though it is not statistically significant.}
    \label{fig:sports_baseline}
\end{figure}

\subsubsection{Results}

Results for the pairwise comparison with external models and controlled baselines are shown in \autoref{fig:sports_rq1} and \autoref{fig:sports_baseline}, respectively.

\paragraph{External models} As shown in \autoref{fig:sports_rq1}, we find that \name outperforms all external models on this sports data in prompt alignment, real-world fidelity, and overall holistic preference when evaluated by external annotators.
For visual quality, we find that \name surpasses Cosmos2 variants, has similar performance as MAGI-1 variants, and is worse than WAN 2.2 5B. For Cosmos2, this is expected, as this model was largely tuned for performance on domains like robotics and autonomous driving. For more general domain pretrained models like MAGI-1 and WAN 2.2, sports provides a challenging test case.

\paragraph{Controlled baselines} In \autoref{fig:sports_baseline}, the \name model shows improvements on holistic preference against both baselines: \name videos have a win-rate 19 points higher than \TtoV (54\% \textit{vs.} 35\%, with 11\% ties) and 12 points higher than \ThinktoV (53\% \textit{vs.} 41\%, with 6\% ties), although the latter is not statistically significant.
Furthermore, for alignment, \name has a win-rate 20 points higher than \TtoV and 12 points higher than \ThinktoV (although again, the latter is not statistically significant).
% Holistically, \name videos are preferred 57.1\% of the time relative to \ThinktoV videos (with 22.6\% ties) and 61.7\% of the time relative to \TtoV (with 21.7\% ties). 
\name shows similar real world fidelity and visual quality compared to \TtoV and \ThinktoV.

\medskip
Together, these results demonstrate that strong generation performance in real-world, dynamic contexts can be achieved with the \name method.

\subsubsection{Qualitative analysis}

\begin{table*}[t]
\centering
\begin{tabular}{@{}lp{0.85\textwidth}@{}}
\toprule
\multicolumn{2}{p{0.99\textwidth}}{\textbf{Meta-prompt:} \emph{A player in a white jersey dribbles past defenders, kicks the ball towards the goal, and scores, celebrating as the goalkeeper remains on the ground.}} \\
\midrule
\multicolumn{2}{l}{\textbf{Generated video (conditioned on 1 frame):}} \\[4pt]
\multicolumn{2}{c}{%
\begin{tabular}{@{}*{8}{c}@{}}
\includegraphics[width=0.10\textwidth]{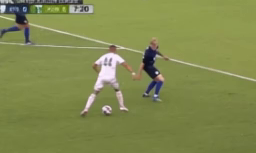} &
\includegraphics[width=0.10\textwidth]{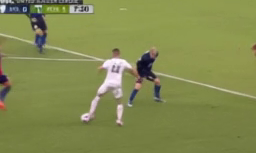} &
\includegraphics[width=0.10\textwidth]{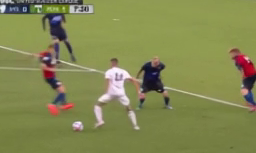} &
\includegraphics[width=0.10\textwidth]{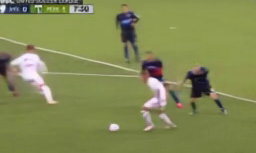} &
\includegraphics[width=0.10\textwidth]{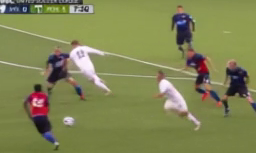} &
\includegraphics[width=0.10\textwidth]{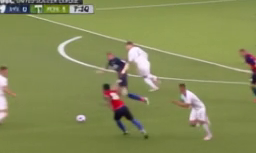} &
\includegraphics[width=0.10\textwidth]{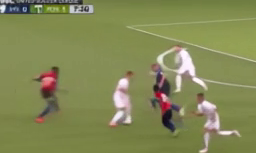} &
\includegraphics[width=0.10\textwidth]{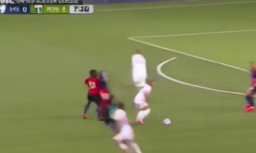} \\[4pt]
\includegraphics[width=0.10\textwidth]{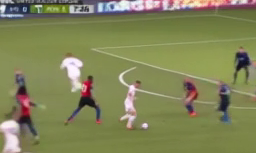} &
\includegraphics[width=0.10\textwidth]{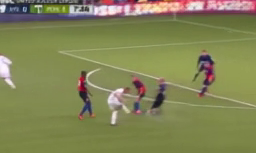} &
\includegraphics[width=0.10\textwidth]{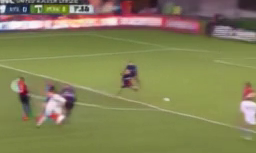} &
\includegraphics[width=0.10\textwidth]{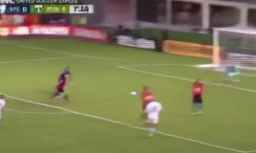} &
\includegraphics[width=0.10\textwidth]{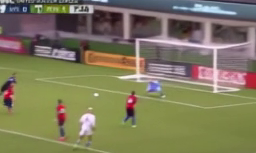} &
\includegraphics[width=0.10\textwidth]{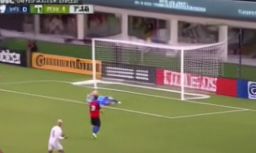} &
\includegraphics[width=0.10\textwidth]{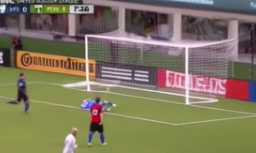} &
\includegraphics[width=0.10\textwidth]{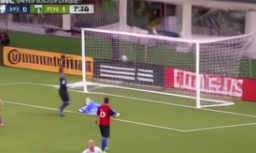}
\end{tabular}
} \\[8pt]
\midrule
\textbf{Timestamp} & \textbf{Generated interleaved text during video generation} \\
\midrule
t=0.0 & {A soccer player in a white jersey dribbles the ball while being closely defended by two players in red jerseys. The player in white maintains possession, maneuvering around the defenders. Another player in a red jersey is positioned further back, observing the play. The action takes place on a grass field with a crowd of spectators in the background.} \\[4pt]
t=0.56 & {The player in the white jersey kicks the ball forward, and a teammate in a white jersey runs towards it. The defenders in red jerseys chase after the ball, with one attempting to intercept. The player in the white jersey gains possession and continues dribbling.} \\[4pt]
t=2.56 & {The player in the white jersey kicks the ball towards the goal. The goalkeeper dives to the left in an attempt to save it.} \\[4pt]
t=4.63 & {The player in the white jersey celebrates the goal by running towards the right side of the frame. The goalkeeper remains on the ground.} \\
\bottomrule
\end{tabular}
\caption{\textbf{\name interleaved text and video generation rollout.} 16 subsampled frames illustrate the evolution of the scene with model-generated interleaved planning. Each generated text helps plan for the video segment following its timestamp.}
\label{tab:sports_rq1_frames}
\end{table*}

\begin{table*}[t]
\centering
\begin{tabular}{@{}lp{0.85\textwidth}@{}}
\toprule
\multicolumn{2}{p{0.99\textwidth}}{\textbf{Meta-prompt:} \emph{A player in a white jersey dribbles past a defender in an orange jersey, shoots, and scores. The players then walk away from the hoop.}} \\
\midrule
\multicolumn{2}{l}{\textbf{Generated video (conditioned on 1 frame):}} \\[4pt]
\multicolumn{2}{c}{%
\begin{tabular}{@{}*{8}{c}@{}}
\includegraphics[width=0.10\textwidth]{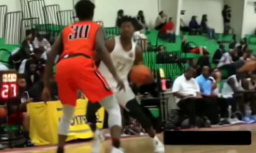} &
\includegraphics[width=0.10\textwidth]{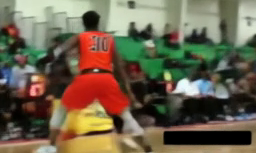} &
\includegraphics[width=0.10\textwidth]{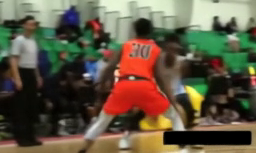} &
\includegraphics[width=0.10\textwidth]{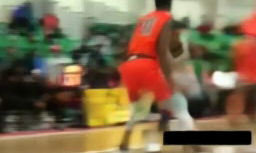} &
\includegraphics[width=0.10\textwidth]{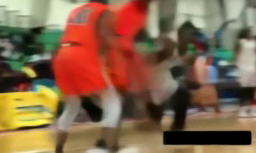} &
\includegraphics[width=0.10\textwidth]{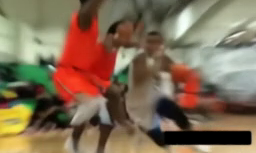} &
\includegraphics[width=0.10\textwidth]{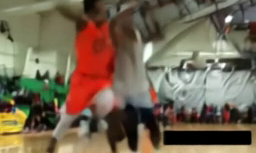} &
\includegraphics[width=0.10\textwidth]{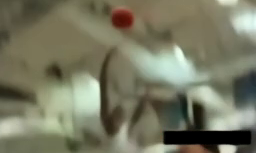} \\[4pt]
\includegraphics[width=0.10\textwidth]{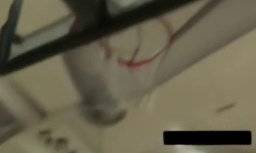} &
\includegraphics[width=0.10\textwidth]{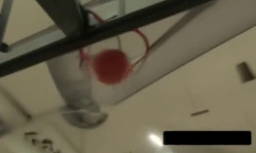} &
\includegraphics[width=0.10\textwidth]{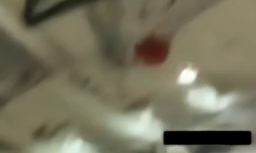} &
\includegraphics[width=0.10\textwidth]{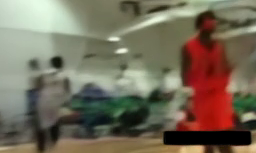} &
\includegraphics[width=0.10\textwidth]{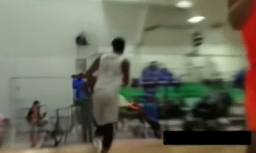} &
\includegraphics[width=0.10\textwidth]{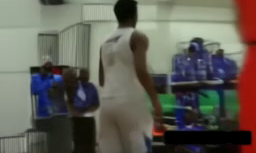} &
\includegraphics[width=0.10\textwidth]{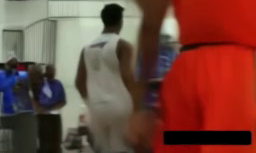} &
\includegraphics[width=0.10\textwidth]{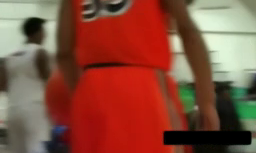}
\end{tabular}
} \\[8pt]
\midrule
\textbf{Timestamp} & \textbf{Generated interleaved text during video generation} \\
\midrule
t=0.0 & {A player in a white jersey dribbles the basketball while being guarded by a player in an orange jersey. The player in white maneuvers around the defender, maintaining control of the ball as they move across the court.} \\[4pt]
t=2.56 & {The player in the white jersey jumps and shoots the basketball toward the hoop. The ball arcs through the air and goes through the net, scoring a basket. The camera follows the ball's trajectory.} \\[4pt]
t=4.88 & {Players in white and orange jerseys walk across the court. The player in the white jersey with the number 1 walks toward the sideline, while the player in the orange jersey with the number 33 walks in the opposite direction.} \\
\bottomrule
\end{tabular}
\caption{\textbf{\name interleaved text and video generation rollout.} 16 subsampled frames illustrate the evolution of the scene with model-generated interleaved planning. Each generated text helps plan for the video segment following its timestamp.}
\label{tab:sports_rq1_frames_b}
\end{table*}

\begin{table*}[t]
\centering
\begin{tabular}{@{}lp{0.85\textwidth}@{}}
\toprule
\multicolumn{2}{p{0.99\textwidth}}{\textbf{Meta-prompt:} \emph{A weightlifter in a blue uniform lifts a barbell from the ground to his shoulders, then raises it above his head, holding it steady. An observer in a blue jacket watches closely.}} \\
\midrule
\multicolumn{2}{l}{\textbf{Generated video (conditioned on 1 frame):}} \\[4pt]
\multicolumn{2}{c}{%
\begin{tabular}{@{}*{8}{c}@{}}
\includegraphics[width=0.10\textwidth]{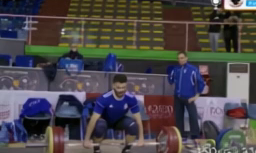} &
\includegraphics[width=0.10\textwidth]{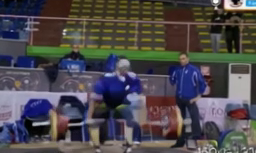} &
\includegraphics[width=0.10\textwidth]{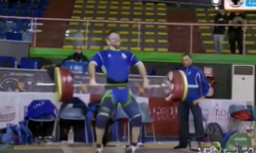} &
\includegraphics[width=0.10\textwidth]{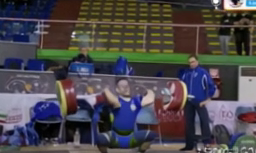} &
\includegraphics[width=0.10\textwidth]{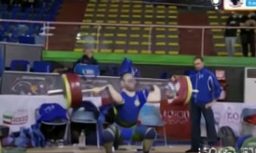} &
\includegraphics[width=0.10\textwidth]{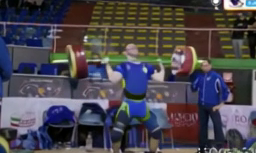} &
\includegraphics[width=0.10\textwidth]{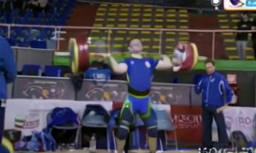} &
\includegraphics[width=0.10\textwidth]{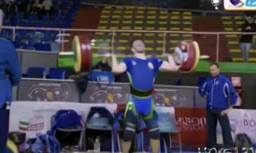} \\[4pt]
\includegraphics[width=0.10\textwidth]{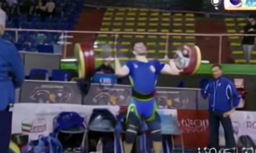} &
\includegraphics[width=0.10\textwidth]{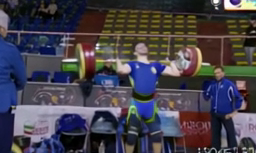} &
\includegraphics[width=0.10\textwidth]{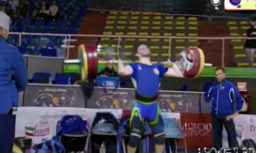} &
\includegraphics[width=0.10\textwidth]{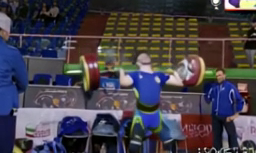} &
\includegraphics[width=0.10\textwidth]{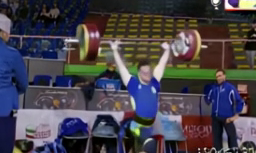} &
\includegraphics[width=0.10\textwidth]{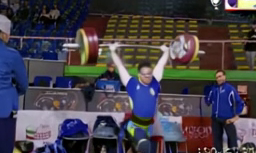} &
\includegraphics[width=0.10\textwidth]{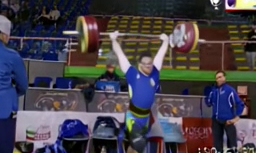} &
\includegraphics[width=0.10\textwidth]{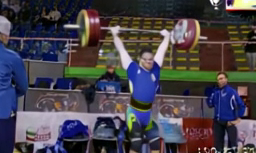}
\end{tabular}
} \\[8pt]
\midrule
\textbf{Timestamp} & \textbf{Generated interleaved text during video generation} \\
\midrule

t=0.0 & {A weightlifter in a blue uniform lifts a barbell from the ground to his shoulders in a swift motion, while another man in a blue jacket watches from the side.} \\[4pt]

t=1.56 & {The weightlifter holds the barbell at shoulder height, standing upright with a focused expression, while the observer remains stationary, watching the lifter's form.} \\[4pt]

t=3.13 & {The weightlifter raises the barbell above his head, fully extending his arms, maintaining a stable stance with legs slightly apart, as the observer continues to watch.} \\[4pt]

t=5.13 & {The weightlifter holds the barbell overhead with arms fully extended, standing still with a steady posture, while the observer remains in place, observing the lifter's position.} \\

\bottomrule
\end{tabular}
\caption{\textbf{\name interleaved text and video generation rollout.} 16 subsampled frames illustrate the evolution of the scene with model-generated interleaved planning. Each generated text helps plan for the video segment following its timestamp.}
\label{tab:sports_rq1_frames_c}
\end{table*}

We provide qualitative examples of generations from \name conditioned on a meta-prompt and one frame. \autoref{tab:sports_rq1_frames}, \autoref{tab:sports_rq1_frames_b}, and \autoref{tab:sports_rq1_frames_c} illustrate the evolution of the scene with model-generated interleaved textual planning. Each generated segment of text helps plan for the subsequent set of frames.  
%Additionally, annotators note that the external models often miss primary actions included in the prompt, and the generated videos tend to contain more erratic movements and less physical realism compared to the \name model.

A key advantage of \name is its controllability through interleaved text: users can inspect, edit, or steer the intermediate textual plan to modify video generation at any timestep. We demonstrate this capability using the 8B-MoT \name model trained on sports data, providing qualitative examples of how interleaved captions dynamically alter generation trajectories. See \autoref{tab:caption_steering}.

\subsection{Discussion}

We demonstrated that co-generating text and video in an interleaved sequence provides advantages over non-interleaved approaches such as \TtoV and \ThinktoV. The gains in gaming (\CSGO) are more pronounced than in sports (\autoref{sec:findings}, \autoref{fig:rq1_results} vs. \autoref{sec:soccer}, \autoref{fig:sports_baseline}). 
Two factors potentially contribute to this gap. First, the density of interleaved text: \CSGO provides frame-level textual signals, while sports rely on synthetic captions inserted every 1.9 seconds on average. Second, the quality of the interleaved text: \CSGO uses ground-truth actions, whereas sports captions are generated by VLMs and often contain hallucinations. 
Despite these challenges, \name scales effectively to real-world data and still outperforms non-interleaved baselines. 

In addition, \name naturally supports user interventions at any point during generation, and its interleaved autoregressive setup allows simple extension to longer videos through sliding windows. 
Overall, interleaved text helps the model learn video generation more effectively and provides flexible control at inference time. Future work could focus on improving the granularity and accuracy of the interleaved text in the training data across more video domains.

%% file: sections/06_relatedwork.tex
\section{Related Work}

\paragraph{Unified multimodal architectures.} The field has rapidly progressed toward unified foundation models capable of understanding and generating content across multiple modalities. Early unified models like Flamingo \citep{alayrac2022flamingo} pioneered this area, integrating vision and language components via cross-attention. Emu2 \citep{sun2023emu} advanced this paradigm by employing a unified autoregressive (AR) objective for both modalities. More recently, Chameleon \citep{chameleonteam2024chameleonmixedmodalearlyfusionfoundation} adopted an early-fusion, purely AR architecture that unifies images and text as discrete tokens, demonstrating flexible multi-modal reasoning.

Subsequent work has focused on improving scalability using hybrid autoregressive and diffusion approaches \citep{zhao2024monoformer, transfusion, showo, metamorph, janusflow, lmfusion}. For instance, Transfusion \citep{transfusion} improved upon Chameleon by leveraging modality-specific losses, using AR for text and diffusion for images, which allows for the use of continuous visual tokens and scales better. Similarly, the Janus series collectively introduced decoupled visual pathways for encoding and generation \citep{janusflow, januspro} to improve scalability while preserving shared semantics. BAGEL \citep{deng2025bagel} scales multimodal pretraining using a Mixture-of-Transformers (MoT) \citep{liang2024mixture}. Manzano \citep{li2025manzano} advances this line by unifying image, text, and video understanding under a coherent generative framework, producing seamless multimodal interleaving. Here, we propose a scalable recipe to extend these works to the generation of video and interleaved multimodal sequences.

\paragraph{Action-conditioned video generation.} Video generation models are increasingly capable of simulating diverse environments~\citep{bruce2024genie,videoworldsimulators2024,xiang2024pandora,teng2025magi,ye2025yan}. The first successes of action-conditioned video generation have been in computer games~\citep{diamond,yu2025gamefactory,hunyuanworld}; for example, GameNGen successfully simulates Doom~\citep{gamengen}. However, models focusing on computer games remain limited to discrete, task-specific actions. Other models extend beyond games to visually diverse environments with physical action spaces, such as navigation~\citep{bar2025navigation,genex} or full-body control~\citep{bai2025whole}. Building a truly general world model requires a flexible and expressive action space. Genie~\citep{bruce2024genie} addresses this by learning latent actions, but interpreting and self-generating such actions remains challenging. In contrast, we aim to build models which can both generate \emph{and} condition video generation on open-ended actions with a vocabulary grounded in text, spanning diverse environments including computer games and sports videos.

\paragraph{Autoregressive video generation.} Modeling long, sequential multimodal inputs, particularly video, remains a critical challenge. Several approaches take an autoregressive approach to iteratively predict the next frame or block of frames conditioned on previous frames \citep{yin2024causvid,chen2025diffusion, agarwal2025cosmos,hacohen2024ltx,lin2025autoregressive,cheng2025playing,yuan2025lumos}, with notable work like MAGI-1 \citep{teng2025magi} focusing on high-fidelity AR video generation. However, these works, including MAGI-1, Cosmos~\citep{agarwal2025cosmos}, and \citet{chen2025diffusion}, primarily focus on video generation and cannot generate interleaved text. Furthermore, Cosmos does not condition on the entire history, limiting long-term coherence. Meanwhile, token-based autoregressive generation—popularized by large language models (LLMs)—is utilized by models like VideoPoet \citep{videopoet}, which treats video generation as a next-token prediction task and can straightforwardly leverage KV caching to improve generation efficiency.

To address the computational intensity and mitigate exposure bias during long-sequence generation, optimization techniques like diffusion forcing \citep{chen2025diffusion} and self-forcing \citep{huang2025selfforcingbridgingtraintest} are employed. Specifically, diffusion forcing introduced progressive noise to parallelize denoising across frames. To alleviate exposure bias in sequential training, a popular strategy is to inject a small amount of noise into clean data \citep{teng2025magi}. 
Our teacher-forcing setup that separately encodes noisy and clean data representations is also adopted in prior image generation work \citep{hu2024acdit}. 

\paragraph{Full-sequence diffusion models.} Another line of work \citep{opensora, opensora2, wan2025} follows the full-sequence paradigm where the video is treated as one large tensor and all frames are denoised simultaneously. However, this results in prohibitively large attention computations, making development of such methods prohibitive for long-form and interactive generation.

\paragraph{Multi-prompt video generation.} A complementary line of work involves extending video generation length using multiple prompts \citep{villegas2023phenaki,  oh2024mevg, bansal2024talc, cai2025ditctrl}. The interleaved design of \name allows for a natural extension of video length by simply using a sliding window approach: additional prompts for continuing the generation can be generated by the model (or inserted by the user) at any timestep.

\begin{table*}[t]
\centering
\begin{tabular}{@{}*{7}{c}@{}}
\toprule
& \multicolumn{6}{c}{\textbf{Subsampled frames}} \\
\midrule
 & t=0.0 & t=1.19 & t=2.38 & t=3.62 & t=4.81 & t=6.06 \\
\cmidrule(lr){2-7}
\multicolumn{7}{p{0.95\textwidth}}{\textbf{Intervention 1 at t=1.56:} {\small The man completes his golf swing, raising the club above his shoulder as he follows through. The golf ball is no longer visible, having been struck and sent flying forward. His body turns slightly to the left, maintaining balance after the powerful swing.}} \\
& \includegraphics[width=0.14\textwidth]{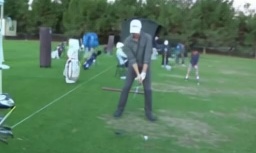} & 
  \includegraphics[width=0.14\textwidth]{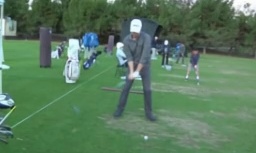} & 
  \includegraphics[width=0.14\textwidth]{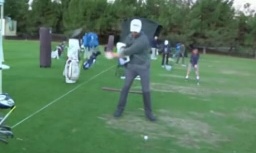} & 
  \includegraphics[width=0.14\textwidth]{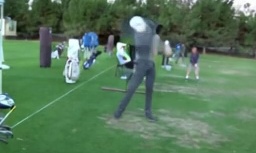} &
  \includegraphics[width=0.14\textwidth]{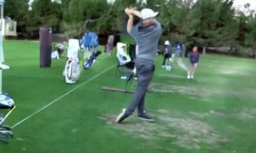} & 
  \includegraphics[width=0.14\textwidth]{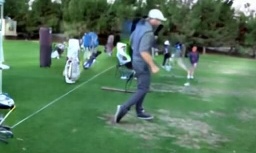} \\
\cmidrule(lr){2-7}
\multicolumn{7}{p{0.95\textwidth}}{\textbf{Intervention 2 at t=1.56:} {\small The man completes his golf swing, raising the club above his shoulder as he follows through. The camera pans to track the ball as it soars through the air.}} \\
& \includegraphics[width=0.14\textwidth]{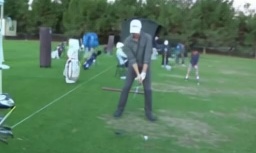} & 
  \includegraphics[width=0.14\textwidth]{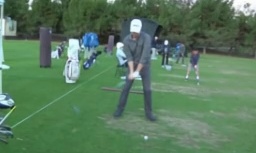} & 
  \includegraphics[width=0.14\textwidth]{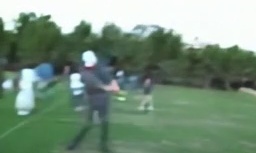} & 
  \includegraphics[width=0.14\textwidth]{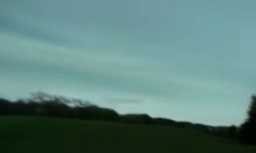} &
  \includegraphics[width=0.14\textwidth]{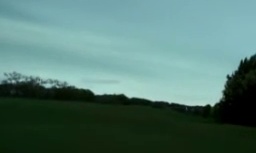} & 
  \includegraphics[width=0.14\textwidth]{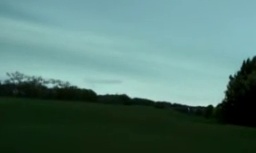} \\
\midrule
 & t=0.0 & t=0.75 & t=1.5 & t=2.31 & t=3.06 & t=3.88 \\
\cmidrule(lr){2-7}
\multicolumn{7}{p{0.95\textwidth}}{\textbf{Intervention 1 at t=1.56:} {\small The player in the white jersey takes control of the ball and runs towards the goal. He kicks the ball powerfully towards the net. The goalkeeper in the yellow jersey leaps to the left, extending his arms in an attempt to block the shot. The ball moves swiftly towards the goalpost as the goalkeeper's jump reaches its peak.}} \\
& \includegraphics[width=0.14\textwidth]{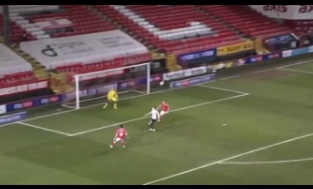} & 
  \includegraphics[width=0.14\textwidth]{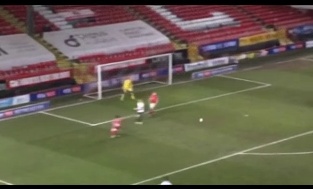} & 
  \includegraphics[width=0.14\textwidth]{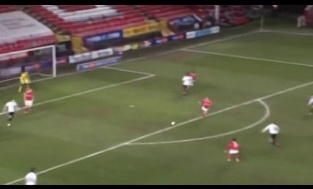} & 
  \includegraphics[width=0.14\textwidth]{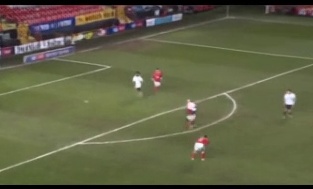} &
  \includegraphics[width=0.14\textwidth]{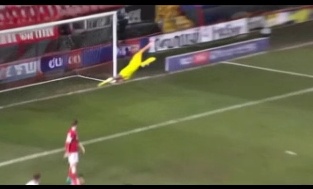} & 
  \includegraphics[width=0.14\textwidth]{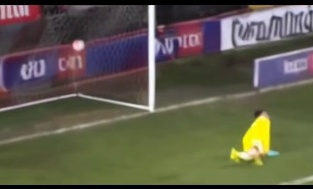} \\
\cmidrule(lr){2-7}
\multicolumn{7}{p{0.95\textwidth}}{\textbf{Intervention 2 at t=1.56:} {\small The player in the red jersey intercepts the ball near the center of the field and starts dribbling towards the right side of the frame, moving away from the goal. He evades an approaching defender in a white jersey by maneuvering the ball skillfully. As he advances, other players adjust their positions, preparing for the next phase of play. The goalkeeper remains near the goal, observing the developing action.}} \\
& \includegraphics[width=0.14\textwidth]{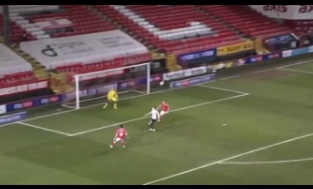} & 
  \includegraphics[width=0.14\textwidth]{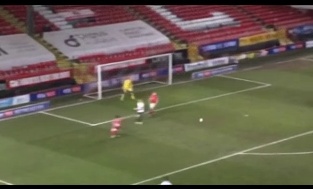} & 
  \includegraphics[width=0.14\textwidth]{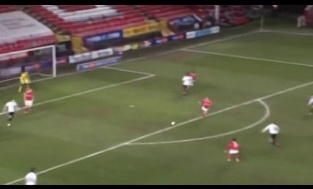} & 
  \includegraphics[width=0.14\textwidth]{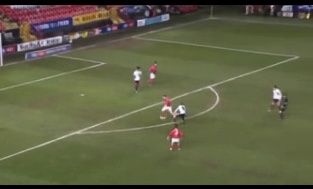} &
  \includegraphics[width=0.14\textwidth]{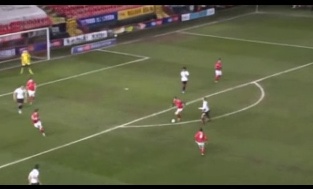} & 
  \includegraphics[width=0.14\textwidth]{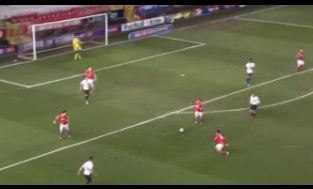} \\
\bottomrule
\end{tabular}
\caption{\textbf{Comparison of video rollouts steered by different interleaved captions.} We alter the interleaved caption at second 1.56s and observe how the generation trajectory is altered.
}
\label{tab:caption_steering}
\vspace{-0.9em}
\end{table*}

\paragraph{Dense captioning pipelines.}
Recent work has developed dense and temporally aligned video captioning approaches that generate hierarchically structured descriptions over long videos, often combining coarse global summaries with fine-grained local captions \citep{chen2024makes, peng2025patch, chen2025planning}. Our pipeline follows this general recipe but emphasize feature-based adaptive segmentation compared to existing pipelines such as InternVid \citep{wang2023internvid}. On the data selection side, our motion-, face-, and VLM-based filters are conceptually related to large-scale video curation efforts such as OpenVid \citep{nan2024openvid}, UltraVideo \citep{xue2025ultravideo}, VideoUFO \citep{wang2025videoufo}, and Koala-36M \citep{wang2025koala}, but are tailored to the sports domain and explicitly optimized for producing high-quality \textit{interleaved} text–video training corpora. Generating action change descriptions is relevant to differential captioning works such as ShareGPT4Video \citep{chen2024sharegpt4video}, ProgressCaptioner~\citep{xue2025progress}, and CI-VID~\citep{ju2025cividcoherentinterleavedtextvideo}. Finally, the \ThinktoV baseline described in \autoref{sec:soccer} closely resembles the descriptive caption ``upsampling'' approach described in \citet{betker2023improving}, which was used to improve DALLE-3 prompt following.

%% file: sections/07_conclusion_limitations.tex
\section{Conclusion}\label{sec:conclusion}

This work introduces \name, a unified modeling framework that decomposes video generation into an interleaved text and video generation process. By generating both text and video in an interleaved manner, this approach offers two key advantages: (1) it offloads much of the semantic complexity of video generation to the model's text generation components, and (2) it enables more flexible and effective user control during generation. Through controlled experiments on video game data, we demonstrate that \name outperforms baseline models in both video generation quality and controllability. We also show how this paradigm scales to real-world data by building a data augmentation pipeline that uses VLMs to enrich video data with interleaved action descriptions and comparing with established and controlled baselines. We believe \name represents a promising step toward unifying advances in language model planning and reasoning with highly controllable video generation systems into a single generative framework.

\section{Acknowledgements} 
We would like to thank Tariq Berrada, Rohit Girdhar, Sachin Mehta, Hritik Bansal, Mike Lewis, and Adriana Romero Soriano for helpful discussions throughout this project.

%% file: sections/08_appendix.tex
\section{Additional details: Controlled Experiments with Video Game Data}\label{appendix:csgo}

\subsection{Model configuration details}\label{appendix:configuration_csgo}
See \autoref{tab:model_configs} for key training and inference hyperparameters.

\begin{table*}[h!]
\centering
\renewcommand{\arraystretch}{1.3}
\begin{tabular}{cc}
\toprule
\textbf{Layers} & 28 \\
\textbf{Model Dimension} & 3072 \\
\textbf{FFN Dimension} &  8192 \\
\textbf{Attention Heads} & 24 \\
\textbf{Key/Value Heads} & 8 \\
\textbf{Activation Function} & SwiGLU \\
\textbf{Vocabulary Size} & 128K \\
\textbf{Positional Embeddings -- Interleaved Sequence} & 1D RoPE \\
\textbf{Positional Embeddings -- Video Only} & 2D APE\\
\textbf{Training Steps} & 50K \\
\textbf{Batch Size} & 128 \\
\textbf{Learning rate} & 3e-4 \\
\textbf{Max Context Length} & 15360 \\
\textbf{Tokens per Frame Chunk} & 240 \\
\textbf{Timestep $t$} & $\operatorname{logistic}(\mathcal{N}(0, 1.96))$ \\
\textbf{Text Dropout Rate $p_{\text{txt-drop}}$} & 0 \\
\textbf{Clean Video Flip Rate $p_{\text{clean-vid-flip}}$} & 0.5 \\
\textbf{Text Sampling Temperature} & 0.7 \\
\textbf{CFG} & 1.0 \\
\textbf{ODE Sampler} & Euler \\
\textbf{ODE Sampling Steps} & 50 \\
\bottomrule
\end{tabular}
\caption{\textbf{Model configuration details} for \name and baselines for experiments on video game data in \autoref{sec:findings}. All model variants adopt a 3B-MoT Transfusion architecture.}
\label{tab:model_configs}
\end{table*}

\subsection{Human Evaluation Task}
\label{app:csgo_eval}

\paragraph{Pairwise comparisons} Annotators evaluate overall video quality within pairings according to the question:
\begin{quote}
    Which video has better visual quality? Examples of poor visual quality include cloudy/flickering generation quality, jumping through walls, static player movement (e.g. moving very slowly in one direction), and random jumps to elesewhere in the map. Do not penalize ghost-like or translucent characters.

    Answer choices:
    \begin{itemize}
        \item Left has significantly better visual quality.
        \item Left has marginally more visual quality.
        \item Unsure or both seem equally good/bad.
        \item Right has marginally more visual quality.
        \item Right has significantly better visual quality.
    \end{itemize}
\end{quote}

\paragraph{Pointwise comparisons} Annotators evaluate alignment between the generated video and intervention prompt according to the following instruction. In this instruction, `Caption A' refers to the user-controllable intervention prompt included with the generated video, and `START' and `STOP' are assistive visual indicators added posthoc to the generated video corresponding to the intervention timestamp: 
\begin{quote}Please watch the video. Does Caption A correctly reflect the clip shown between `START' and `STOP'? Consider primarily the period shown between START/STOP, although if the caption refers to jumping, reloading, or moving backwards you may also consider the period *immediately* following `STOP'. Please use the rest of the video only for general context.

    Answer choices:
    \begin{itemize}
        \item The caption correctly reflects the video.
        \item The caption does not correctly reflect the video.
        \item Unsure - the player is not actively playing (taken down by enemy and can't move).
        \item Unsure - other (please specify in comments).
\    \end{itemize}
\end{quote}

Additionally, visual quality is evaluated with the question:
\begin{quote}
How is the overall visual quality of the video? Examples of poor visual quality include cloudy/flickering generation quality, jumping through walls, static player movement (e.g. moving very slowly in one direction), and random jumps to elesewhere in the map. Do not penalize ghost-like or translucent characters.

Answer choices:
    \begin{itemize}
        \item The video has strong visual quality.
        \item The video has moderate visual quality.
        \item The video has weak visual quality.
        \item The video has no visual quality.
    \end{itemize}
\end{quote}

\section{Additional details: Scaling \name to Real World Data}\label{appendix:sports}
\subsection{VLM-based quality filtering prompt}\label{appendix:prompt}
\begin{tcolorbox}[colback=green!10!white, colframe=black!80!black, boxrule=0.5pt, arc=2pt, left=2pt, right=2pt, top=2pt, bottom=2pt]
You are a capable model that can determine if the video is high quality or low quality, here are criteria to determine the quality:

1) Video is low quality if it has person talking to camera without any other motion, face in corners is OK. \\
2) Video is low quality if there is jittery motion due to camera movement. \\
3) Video is low quality if there is no motion with blank or static screen or image with just zoom in and zoom out. \\
4) Video is high quality if it has meaningful sports content like highlights of a game being played. \\ 

Now please rate video with a score between 1 and 10, where 1 is low quality and 10 is high quality, return the score in json format e.g.: $\{$`quality\_score': $<$predicted score$> \}$. \\

Here are frames from new video sampled from start, middle and end of video:
\end{tcolorbox}

\subsection{VLM-based differential captioning prompt}\label{appendix:differential_caption_prompt}
\begin{tcolorbox}[colback=green!10!white, colframe=black!80!black, boxrule=0.5pt, arc=2pt, left=2pt, right=2pt, top=2pt, bottom=2pt]
You are a cautious video describer. \\
You will be shown multiple video segments from the same source video, shown in chronological order. \\
Describe what happens in EACH segment separately. \\
DO NOT reference `the video' or `the segment' in your descriptions. \\
DO NOT describe any text in the video. \\
Most important: Describe the actions or movements of the main characters or objects in each segment. \\
DO NOT anticipate future actions; only describe actions that are clearly visible in the current segment. \\
DO NOT repeat descriptions from previous segments. \\
If nothing meaningfully changes in the current segment compared to previous segments, use an empty string "" for the description of the current segment. \\
Keep each description concise (under 50 words). DO NOT hallucinate. \\

Format your output EXACTLY as follows: \\ 
Description of segment 1: [your description here] \\ 
Description of segment 2: [your description here] \\
... \\
Description of segment <N>: [your description here] \\ 
Here are the video segments in order: <VIDEO SEGMENTS>\\
\end{tcolorbox}

\subsection{Model configuration details}\label{appendix:configuration_sports}
See \autoref{tab:model_configs_sports} for key training and inference hyperparameters.

\begin{table*}[h!]
\centering
\renewcommand{\arraystretch}{1.3}
\begin{tabular}{cc}
\toprule
\textbf{Layers} & 32 \\
\textbf{Model Dimension} & 4096 \\
\textbf{FFN Dimension} &  14336 \\
\textbf{Attention Heads} & 32 \\
\textbf{Key/Value Heads} & 8 \\
\textbf{Activation Function} & SwiGLU \\
\textbf{Vocabulary Size} & 128K \\
\textbf{Positional Embeddings -- Interleaved Sequence} & 1D RoPE \\
\textbf{Positional Embeddings -- Video Only} & 2D APE\\
\textbf{Training Steps} & 250K \\
\textbf{Batch Size} & 512 \\
\textbf{Learning Rate} & 3e-4 \\
\textbf{Max Context Length} & 13056 \\
\textbf{Tokens per Frame Chunk} & 240 \\
\textbf{Timestep $t$} & $\operatorname{logistic}(\mathcal{N}(0, 1.96))$ \\
\textbf{Text Dropout Rate $p_{\text{txt-drop}}$} & 0.05 \\
\textbf{Clean Video Flip Rate $p_{\text{clean-vid-flip}}$} & 0.2 \\
\textbf{Text Sampling Temperature} & 0.7 \\
\textbf{CFG} & 7.5 \\
\textbf{ODE Sampler} & Euler \\
\textbf{ODE Sampling Steps} & 50 \\
\bottomrule
\end{tabular}
\caption{\textbf{Model configuration details} for \name and baselines for experiments on real world sports data in \autoref{sec:soccer}. All model variants adopt an 8B-MoT Transfusion architecture.}
\label{tab:model_configs_sports}
\end{table*}

\subsection{Human Evaluation Task}
\label{appendix:sports_human_eval}

\begin{figure}[h]
    \centering
    \includegraphics[width=0.9\linewidth]{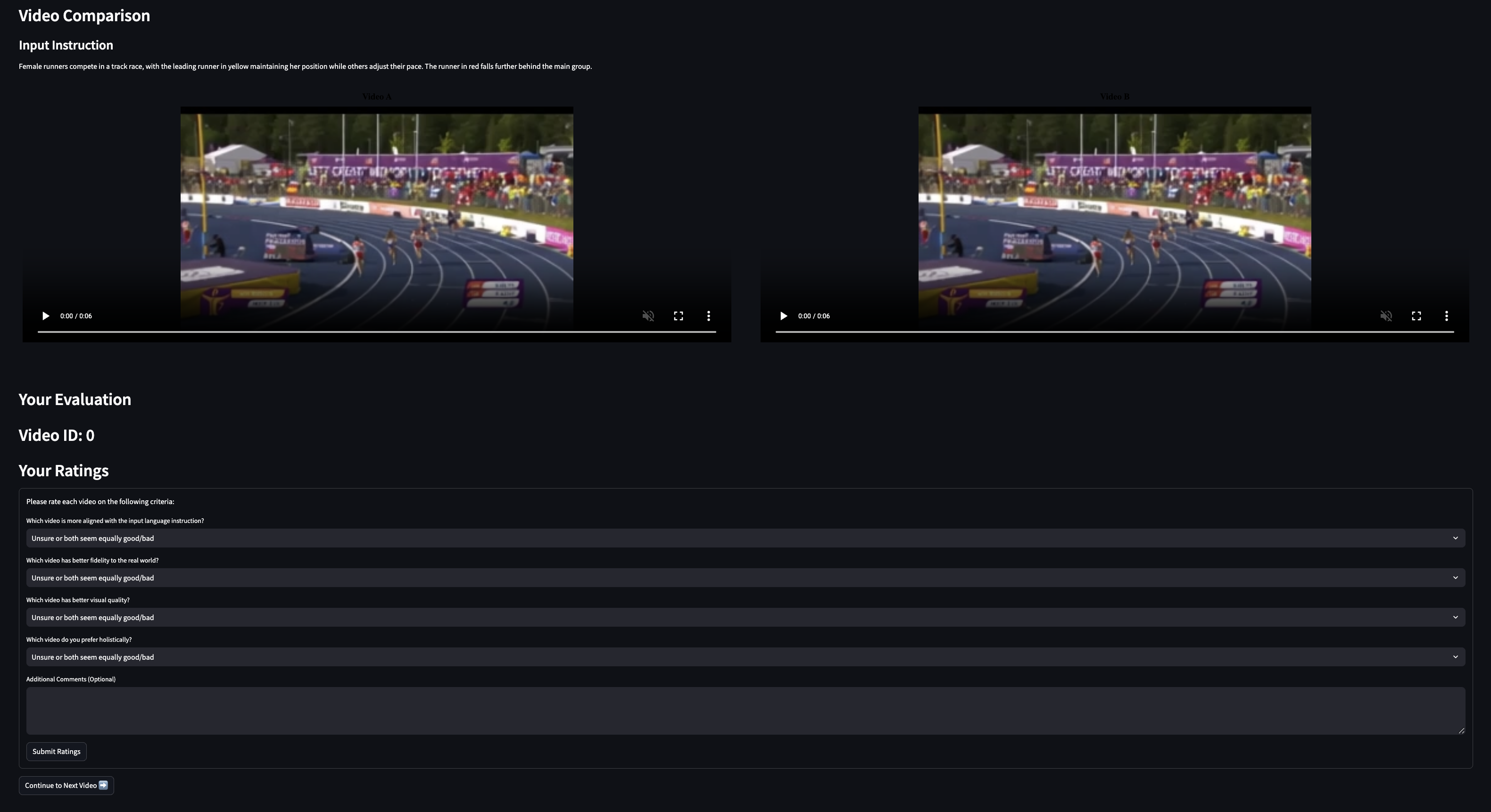}
    \caption{Example UI for evaluating alignment, fidelity, quality, and overall preference in \name and comparison models.}
    \label{fig:ui_example}
\end{figure}

For the human evaluation, all pairs are evaluated by a pool of professional external annotators via the Turing platform for increased robustness. 
A similar user interface as the one used by annotators is shown in Figure~\ref{fig:ui_example}.

We include the evaluation questions answered by the annotators for the results discussed in \autoref{sec:soccer}:

\begin{itemize}
    \item \textbf{Prompt alignment}: Which video is more aligned with the input language instruction?
    \begin{itemize}
        \item Left is significantly more aligned with the text instruction.
        \item Left is marginally more aligned with the text instruction.
        \item Unsure or both seem equally good/bad.
        \item Right is marginally more aligned with the text instruction.
        \item Right is significantly more aligned with the text instruction.
    \end{itemize}
    \item \textbf{Real world fidelity}: Which video has better fidelity to the real world?
    \begin{itemize}
        \item Left has significantly better fidelity to the real world.
        \item Left has marginally better fidelity to the real world.
        \item Unsure or both seem equally good/bad.
        \item Right has marginally better fidelity to the real world.
        \item Right has significantly better fidelity to the real world.
    \end{itemize}
    \item \textbf{Visual quality}: Which video has better visual quality?
    \begin{itemize}
        \item Left has significantly better visual quality.
        \item Left has marginally better visual quality.
        \item Unsure or both seem equally good/bad.
        \item Right has marginally better visual quality.
        \item Right has significantly better visual quality.
    \end{itemize}
    \item \textbf{Holisitic preference}: Which video do you prefer holistically?
    \begin{itemize}
        \item I strongly prefer Left.
        \item I somewhat prefer Left.
        \item Unsure or both seem equally good/bad.
        \item I somewhat prefer Right.
        \item I strongly prefer Right.
    \end{itemize}
    
\end{itemize}